\documentclass{article}

\usepackage{arxiv}

\usepackage[utf8]{inputenc} 
\usepackage[T1]{fontenc}    
\usepackage{hyperref}       
\usepackage{url}            
\usepackage{booktabs}       
\usepackage{amsfonts}       
\usepackage{nicefrac}       
\usepackage{microtype}      
\usepackage{lipsum}         
\usepackage{graphicx}
\usepackage{natbib}
\usepackage{doi}
\usepackage{amsmath,amsfonts}
\usepackage{algorithmic}
\usepackage{algorithm}
\usepackage{array}
\usepackage[caption=false,font=footnotesize,labelfont=rm,textfont=rm]{subfig}
\usepackage{textcomp}
\usepackage{stfloats}
\usepackage{url}
\usepackage{verbatim}
\usepackage{multirow}
\usepackage{xcolor}
\usepackage{amsthm}
\usepackage{cleveref}       
\newtheorem{myDef}{Definition}

\usepackage{enumitem}
\usepackage{makecell}

\title{Fast Adaptive Anti-Jamming Channel Access via Deep Q Learning and Coarse-Grained Spectrum Prediction}

\newif\ifuniqueAffiliation
\uniqueAffiliationtrue

\ifuniqueAffiliation 
\author{ {\hspace{1mm}Jianshu Zhang}\\
	School of Communications and Information Engineering\\
	Nanjing University of Posts and Telecommunications\\
	School of Computer Engineering\\
	Nanjing Institute of Technology\\
	Nanjing, China \\
	\texttt{jianshu.zhang@foxmail.com} \\
	\And
	{\hspace{1mm}Xiaofu Wu} \\
	National Engineering Research Center of Communications and Networking\\
	Nanjing University of Posts and Telecommunications\\
	Nanjing, China \\
	\texttt{xfuwu@ieee.org} \\
	\And
	{\hspace{1mm}Junquan Hu} \\
	College of Communications Engineering\\
	Army Engineering University of PLA\\
	Nanjing, China \\
	\texttt{redmob@sina.com} \\
}
\fi

\renewcommand{\headeright}{}
\renewcommand{\undertitle}{}
\renewcommand{\shorttitle}{}

\hypersetup{
pdftitle={A template for the arxiv style},
pdfsubject={q-bio.NC, q-bio.QM},
pdfauthor={David S.~Hippocampus, Elias D.~Striatum},
pdfkeywords={First keyword, Second keyword, More},
}

\begin{document}
\maketitle

\begin{abstract}
This paper investigates the anti-jamming channel access problem in complex and unknown jamming environments, where the jammer could dynamically adjust its strategies to target different channels. Traditional channel hopping anti-jamming approaches using fixed patterns are ineffective against such dynamic jamming attacks. Although the emerging deep reinforcement learning (DRL) based dynamic channel access approach could achieve the Nash equilibrium (NE) under fast-changing jamming attacks, it requires extensive training episodes. To address this issue, we propose a fast adaptive anti-jamming channel access approach guided by the intuition of ``learning faster than the jammer", where a synchronously updated coarse-grained spectrum prediction serves as an auxiliary task for the deep Q network (DQN) based anti-jamming model. This helps the model identify a superior Q-function compared to standard DRL while significantly reducing the number of training episodes. Numerical results indicate that the proposed approach significantly accelerates the rate of convergence in model training, reducing the required training episodes by up to 70\% compared to standard DRL. Additionally, it also achieves a 10\% improvement in throughput over NE strategies, owing to the effective use of coarse-grained spectrum prediction.
\end{abstract}


\makeatletter
\renewcommand\@makefnmark{}
\makeatother
\footnotetext{This work has been submitted to the IEEE for possible publication. Copyright may be transferred without notice, after which this version may no longer be accessible.}

\section{Introduction}
Wireless communications have found extensive applications in both civilian and military scenarios, where ensuring robust anti-jamming capabilities is paramount for secure transmissions, given the vulnerability of wireless links \cite{8999433, 8387816}. The dynamic characteristics of wireless channels present significant challenges for conventional model-based anti-jamming methods \cite{8412128, 10107729, 8586930}. It has been shown that the model-free approaches, such as reinforcement learning (RL) \cite{watkins1992q, mnih2015human}, could help to empower the legitimate user with enhanced anti-jamming capability in wireless communication environments that are dynamic and unknown \cite{8314744, 9136780}. However, jammers have also evolved intelligently with advancements in Universal Software Radio Peripheral (USRP) \cite{7562476} and artificial intelligence (AI) \cite{7362035} in recent years. Significant breakthroughs have recently been made in the field of smart jamming and intelligent jamming \cite{8958575, 9861743, 9316277, 10279385, 9878270}.

Considerable efforts have been devoted to combating intelligent jamming attacks. In \cite{9449830}, the authors treated the jammer as an integral component of the environment and introduced a DRL based algorithm for combating an RL-based jammer. Xu \cite{9089307} proposed a Deep Recurrent Q-Network (DRQN) based channel access method to identify an optimal access policy that minimizes conflict probability and simultaneously maximizes channel utilization in multi-user scenarios. Additionally, the DRL-based algorithm has potential applications in underwater anti-jamming communications \cite{8254362} and unmanned aerial vehicles (UAV) anti-jamming communications \cite{9992013}. Li \cite{9989422} proposed a Markov decision process (MDP) with a two-dimensional action space (transmit frequency and power) and introduced a dual action network based DRL algorithm with an action feedback mechanism to address the anti-jamming problem in frequency hopping (FH) communication systems. A parallel policy network-based DRL algorithm, which adapts transmission power while accessing idle channels, was introduced in \cite{9751039} to address the anti-jamming MDP problem. In \cite{10227374}, a labeled DRL-based dynamic spectrum anti-jamming approach was proposed to adapt to the fast-changing jamming environment. Considering the non-stationarity characteristic of the environment where the jammer could also be regarded as an RL agent \cite{9733393}, several works have formulated the interaction between the legitimate user and the intelligent jammer within a game theory framework. Xiao \cite{8412128} proposed a Two-Dimensional anti-jamming communication scheme using a hotbooting deep Q-network to enhance mobile device utility and signal quality under cooperative jamming attacks. In \cite{10460356}, Zhang explored the interactions between the legitimate user with faking-slot transmission and the intelligent reactive jammer within a bi-matrix game framework and derived the equilibrium for the game using the quadratic programming method. Additionally, a Neural Fictitious Self-Play (NFSP) method was proposed in \cite{9775208} to identify the approximate NE solution for the dynamic game of radar anti-jamming with imperfect information by employing Minimax Q network \cite{9292435}, and a deceptively adversarial attack approach was proposed in \cite{9500773} to tackle the challenge presented by smart jamming.

On the other hand, modeling the interaction between players using multi-agent reinforcement learning and leveraging the behaviors of other agents in multi-agent systems has gained considerable attention in recent years \cite{20164903082829, 20182105216256, 20184205945705, 20182105216196, 20232414251606, 20232614296188}. The use of deep reinforcement opponent network (DRON) \cite{20164903082829} could achieve superior performance over that of DQN and its variants in multi-agent environments. Moreover, the model-based opponent modeling (MBOM) \cite{20232614296188} could simulate the iterative reasoning process within the environment model and generate a range of opponent policies for achieving a more effective adaptation in a variety of tasks. To address the anti-jamming problems in wireless communications, Li \cite{9939159} proposed an opponent modeling based anti-intelligent jamming (OMAIJ) algorithm that analyzes the jammer's policy and targets its vulnerabilities. Yuan \cite{10308595} introduced an opponent awareness-based anti-jamming algorithm that considers the jammer's learning to effectively counter intelligent jamming attacks.

Although the aforementioned works achieve additional performance gains by opponent modeling, their direct application in practical anti-jamming scenarios may encounter significant challenges. This is because these methods are based on the assumption that the user has complete knowledge about the jammer's action space and even its action at each step, which is not always reasonable in real-world scenarios. Additionally, the widely used $\epsilon$-greedy method for exploring \cite{9322486, 9366492} and the inefficient practice of testing a single action per step \cite{9105045} may result in the slow convergence in RL-based methods. If the RL-based anti-jamming approach fails to converge before changes occur in the jammer's strategy, its effectiveness may be significantly diminished \cite{10227374, 10147237, 9264659}.

In this paper, a novel fast adaptive channel access approach, which combines DQN and coarse-grained spectrum prediction, is proposed for anti-jamming. Unlike previous works that focus exclusively on reinforcement learning or opponent modeling, we propose using coarse-grained spectrum prediction as an auxiliary task to accelerate the rate of convergence during the training of the DQN-based anti-jamming model and identify a superior Q-function compared to standard DRL. This integration of spectrum prediction with DQN represents a key innovation, enabling the model to identify fast adaptive anti-jamming strategies more effectively and efficiently, which is not addressed in prior studies. The proposed approach exhibits a faster rate of convergence than both the DRL-based approach and the opponent modeling approach during model training, outperforming the Nash equilibrium in scenarios involving DRL-based jammers. We believe these contributions provide a significant improvement over existing anti-jamming methods, addressing the limitations of the slow rate of convergence during training and the lack of adaptability to dynamic jamming strategies seen in earlier work. The contributions of this paper are outlined below.
\begin{itemize}
\item{Firstly, we describe the adversarial scenario in which a fixed-mode jammer and a DRL-based jammer operate simultaneously. The interaction between the legitimate user and the DRL-based jammer could be formulated as a Markov Game (MG), where the user and the jammer have completely opposing objectives.}
\item{Furthermore, we introduce a novel fast adaptive anti-jamming channel access approach, with coarse-grained spectrum prediction serving as an auxiliary task for the DQN-based anti-jamming model, to identify the dynamic best response to the jammer with time-varying strategies.}
\item{Finally, the advantage of the proposed approach over several existing DRL-based approaches and opponent modeling approaches with respect to the anti-jamming performance as well as the rate of convergence in model training is demonstrated via simulations.}
\end{itemize}

The remainder of this paper is organized as follows. The interaction between the legitimate user and the jammers is formulated in Section \uppercase\expandafter{\romannumeral2}. Subsequently, a supervised learning based coarse-grained spectrum prediction scheme is introduced in Section \uppercase\expandafter{\romannumeral3}, followed by a novel fast adaptive anti-jamming channel access approach with joint DQN and coarse-grained spectrum prediction in Section \uppercase\expandafter{\romannumeral4}. Simulation results regarding the anti-jamming performance and the rate of convergence rate in model training of the proposed approach are provided in Section \uppercase\expandafter{\romannumeral5}. Section \uppercase\expandafter{\romannumeral6} concludes the paper.

\section{Preliminary}
\subsection{System Model}
\begin{figure}[!htb]
\centering
\includegraphics[width=3.3in]{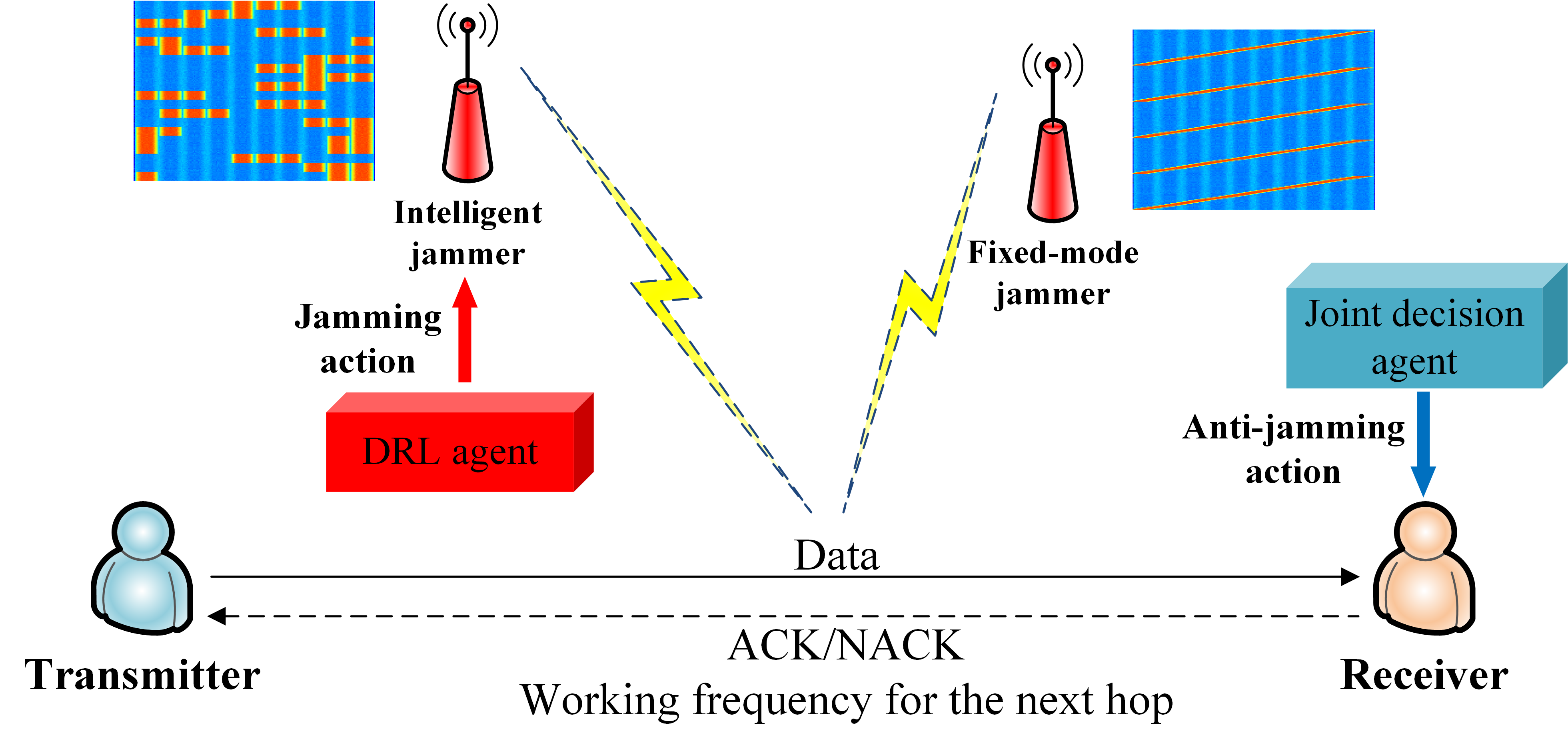}
\caption{System model.}
\label{fig_1}
\end{figure}

As shown in Fig. \ref{fig_1}, a legitimate transmitter communicates with its receiver in the presence of a fixed-mode jammer and an intelligent jammer. The fixed-mode jammer could launch traditional jamming attacks with fixed jamming pattern, e.g., sweeping jamming \cite{9733393}, comb jamming \cite{20075110977801}, partial-band jamming \cite{5473884} etc. Motivated by \cite{8314744}, we consider a DRL-based intelligent jammer that could adaptively adjust its jamming channel to disturb legitimate transmissions.

Assuming that the communication band $\left [ f_L, f_U \right ]$, with total bandwidth $B = f_U - f_L$, could be divided into $M \in \mathbb{Z}^+$ non-overlapping channels. The available channel set is denoted as $\mathcal{F} = \left \{ f_1, f_2, \ldots, f_M \right \}$, where the bandwidth of each channel is $b = B/M$. We consider a synchronous time-slotted system, in which the basic time slots of the legitimate user and the DRL-based intelligent jammer are perfectly aligned. The duration of each basic time slot is $\Delta t$, which is the smallest unit of time resolution. For convenience, the term ``basic time slot'' is abbreviated as ``time slot" in what follows. During the $k$-th time slot (i.e., from $(k-1)\Delta t$ to $k \Delta t$), the legitimate user selects an available channel $f_k^u \in \mathcal{F}$ for transmission. Concurrently, the fixed-mode jammer releases jamming signals on channel $f_k^s$, while the intelligent jammer targets $N_I$ consecutive channels in $\mathcal{I}_k = \left \{ f_k^i \right \}_{i=1, 2, \ldots, N_I}$. This setting reflects the typical contiguous-band emission of practical wideband jammers and establishes a strong adversarial scenario for evaluating the robustness of anti-jamming strategies \cite{9733393}. Similar to \cite{6542754}, the block fading channel model is assumed in the proposed anti-jamming model. Specifically, the channel gain from the transmitter to the receiver during the $k$-th time slot is defined as 
\begin{equation}
h_k^{u, r}(f_k^u) = (d_{u, r})^{-\alpha_d} \xi_k^{f_k^u}, \tag{1}\label{1}
\end{equation}
where $d_{u, r}$, $\alpha_d$, and $\xi_k^{f_k^u}$ denote the distance, the path-loss exponent, and the instantaneous fading coefficient between the legitimate transmitter and the receiver. Similarly, during the $k$-th time slot, the channel gains from the DRL-based jammer to the legitimate receiver is defined as
\begin{equation}
h_k^{i, r}(f_k^i) = (d_{i, r})^{-\alpha_d} \xi_k^{f_k^i}, \tag{2}\label{2}
\end{equation}
and the channel gains from the fixed-mode jammer to the legitimate receiver is defined as
\begin{equation}
h_k^{s, r}(f_k^s) = (d_{s, r})^{-\alpha_d} \xi_k^{f_k^s}. \tag{3}\label{3}
\end{equation}

Then, the power spectral density (PSD) function at the receiver during the $k$-th time slot could be expressed as
\begin{equation}
S_k(f) = h_k^{u, r}(f_k^u) U(f - f_k^u) + h_k^{s, r}(f_k^s) J_s(f - f_k^s) + \sum_{i=1}^{N_I} h_k^{i, r}(f_k^i) J_i(f - f_k^i) + N_k(f), \tag{4}\label{4}
\end{equation}
where $U(f)$ is the PSD function of the legitimate user's baseband signal, $J_i(f)$ is the PSD function of the DRL-based intelligent jammer's baseband signal, $J_s(f)$ is the PSD function of the fixed-mode jammer's baseband signal, and $N_k(f)$ is the PSD function of noise. The legitimate receiver is capable of sensing the entire communication band, and the spectrum vector during the time interval of $\left [ (k-1)\Delta t, k \Delta t \right ]$ (i.e., the $k$-th time slot) could be represented as
\begin{equation}
{\bf{s}}_k = \left ( s_k^1, s_k^2, \ldots, s_k^{N_F} \right ), \tag{5}\label{5}
\end{equation}
with
\begin{align}
s_k^\ell = 10 \log \left [ \int_{(\ell-1) \Delta f}^{\ell \Delta f} S_k(f + f_L) df \right ], \ \ell=1, 2, \ldots, N_F, \tag{6}\label{6}
\end{align}
where the PSD function $S_k(f + f_L)$ could be estimated by P-Welch algorithm \cite{1161901} with the time-domain signals sampled from the $k$-th time slot, and $\Delta f = B/N_F$ is the resolution of spectrum analysis.

Additionally, we assume that each hop has a time duration of $T_h = N_h \Delta t$. To reflect the state of each available channel in a hop, we define the coarse-grained spectrum during the $n$-th hop (i.e., from $(n-1) T_h$ to $n T_h$) as
\begin{equation}
{\bf{c}}_n = \left ( c_n^1, c_n^2, \ldots, c_n^M \right ). \tag{7}\label{7}
\end{equation}
The $m$-th element of the coarse-grained spectrum, i.e., $c_n^m$, represents the discrete spectrum sample value on the $m$-th channel during the $n$-th hop, and could be calculated by
\begin{equation}
c_n^m = \int_{(m-1)\Delta f^\prime}^{m\Delta f^\prime} S_n(f+f_L) df, \quad m=1, 2, \ldots, M, \tag{8}\label{8}
\end{equation}
where $S_n(f+f_L)$ denotes the PSD function estimated over the samples during the $n$-th hop, and $\Delta f^\prime = B/M = b$ is the resolution of coarse-grained spectrum analysis.

\begin{figure}[!t]
\centering
\includegraphics[width=5in]{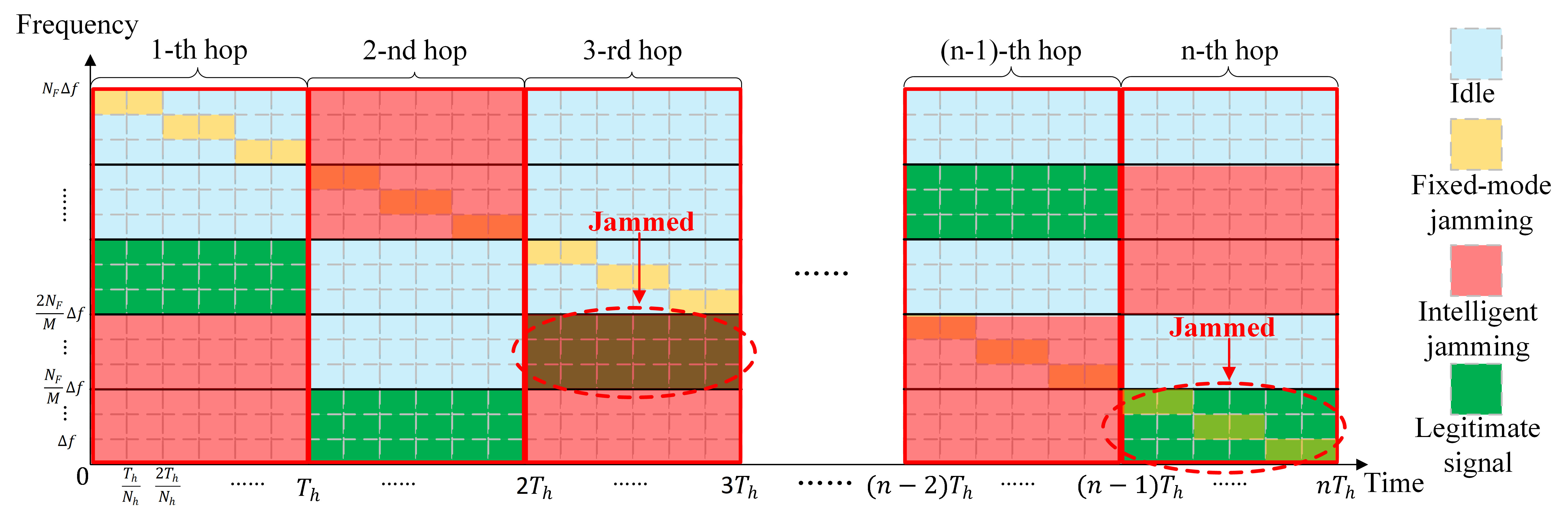}
\caption{An illustrative diagram of the communication time slot structure.}
\label{fig_2}
\end{figure}

\begin{figure}[!t]
\centering
\includegraphics[width=5in]{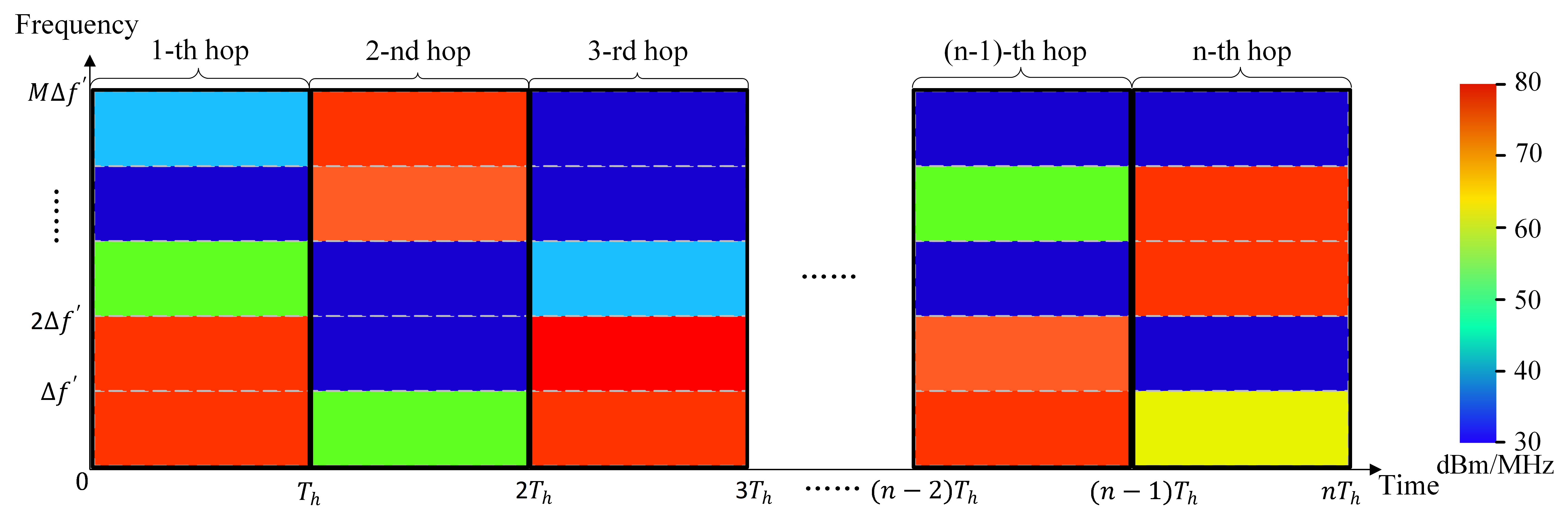}
\caption{The thermodynamic chart of coarse-grained spectrums in several hops.}
\label{fig_3}
\end{figure}

The illustration of the spectrum vectors and the coarse-grained spectra in different hops are shown in Fig. \ref{fig_2} and Fig. \ref{fig_3}, respectively. The small rectangles in Fig. \ref{fig_2} represent the samples of spectrum vectors, and the rectangles in Fig. \ref{fig_3} represent the samples of coarse-grained spectra. The time resolution and the frequency resolution of spectrum analysis are $\Delta t = T_h / N_h$ and $\Delta f = B/N_F$, respectively. The spectrum defined in eq. (\ref{5}) could be employed to learn the behavior of both the legitimate user and the jammers during each time slot. While the time resolution and the frequency resolution of coarse-grained spectrum analysis are $T_h$ and $\Delta f^\prime = B / M = b$, respectively. The coarse-grained spectrum defined in eq. (\ref{7}) is used to express the channel state during each hop with $M$ samples. It is noted that, when representing the spectrum within the same time duration and frequency range, the number of samples in the spectrum matrix $\left [ {\bf{s}}_{(n-1)N_h+1}, \ldots, {\bf{s}}_{nN_h-1}, {\bf{s}}_{nN_h} \right ]^\top$ is $(N_h \times \frac{N_F}{M})$ times greater than the coarse-grained spectrum ${\bf{c}}_n$.

The Signal-to-Interference-plus-Noise Ratio (SINR) at the legitimate receiver defined in eq. (\ref{9}) is often utilized to evaluate the quality of the received signal. If the SINR exceeds the given demodulation threshold $\beta_\text{th}$, the user successfully mitigates jamming attacks. Otherwise, the legitimate transmission fails. The legitimate user aims to find an appropriate channel at the beginning of each hop for achieving a receiving SINR that exceeds $\beta_\text{th}$ during each time slot of the hop.

\begin{equation}
\beta(k, f_u^k) = \frac{h_k^{u, r}(f_k^u) \int_{-b/2}^{b/2} U(f) df}{\int_{f_k^u-b/2}^{f_k^u+b/2} \left [ h_k^{s, r}(f_k^s)J_s(f - f_k^s) + \sum\limits_{i=1}^{N_I} h_k^{i, r}(f_k^i)J_i(f - f_k^i) + N_k(f) \right ] df}. \tag{9}\label{9}
\end{equation}

\subsection{Markov Game Model}
In the proposed anti-jamming scenario, the adaptive channel access decision-making process of the legitimate user is sequential, and the nonstationarity evolution of the state happens when the intelligent jammer adopts a time-varying jamming policy. Under these circumstances, the evolution of the environment is influenced by the actions of both the legitimate transmitter and the intelligent jammer, rather than solely by the action of the legitimate transmitter. This non-stationary characteristic prompts us to model the interaction between the legitimate user and the intelligent jammer as an MG, where the user and the jammer could make decisions simultaneously, each pursuing completely opposing objectives. The anti-jamming MG could be described by a tuple, namely,
\begin{equation}
\mathcal{G} = \left \{ \mathcal{S}, \mathcal{A}^u, \mathcal{A}^j, \mathcal{P}, \mathcal{R}^u, \mathcal{R}^j, \gamma \right \}, \tag{10}\label{10}
\end{equation}
where $\mathcal{S}$ represents the set of environment states, $\mathcal{A}^u$ and $\mathcal{A}^j$ represent the action set of the legitimate user and the intelligent jammer, respectively, $\mathcal{P}$ denotes the transition function, $\mathcal{R}^u$ is the legitimate user's reward function, $\mathcal{R}^j$ is the intelligent jammer's reward function, and $\gamma$ is the discount factor.

\begin{figure}[!htb]
\centering
\includegraphics[width=2in]{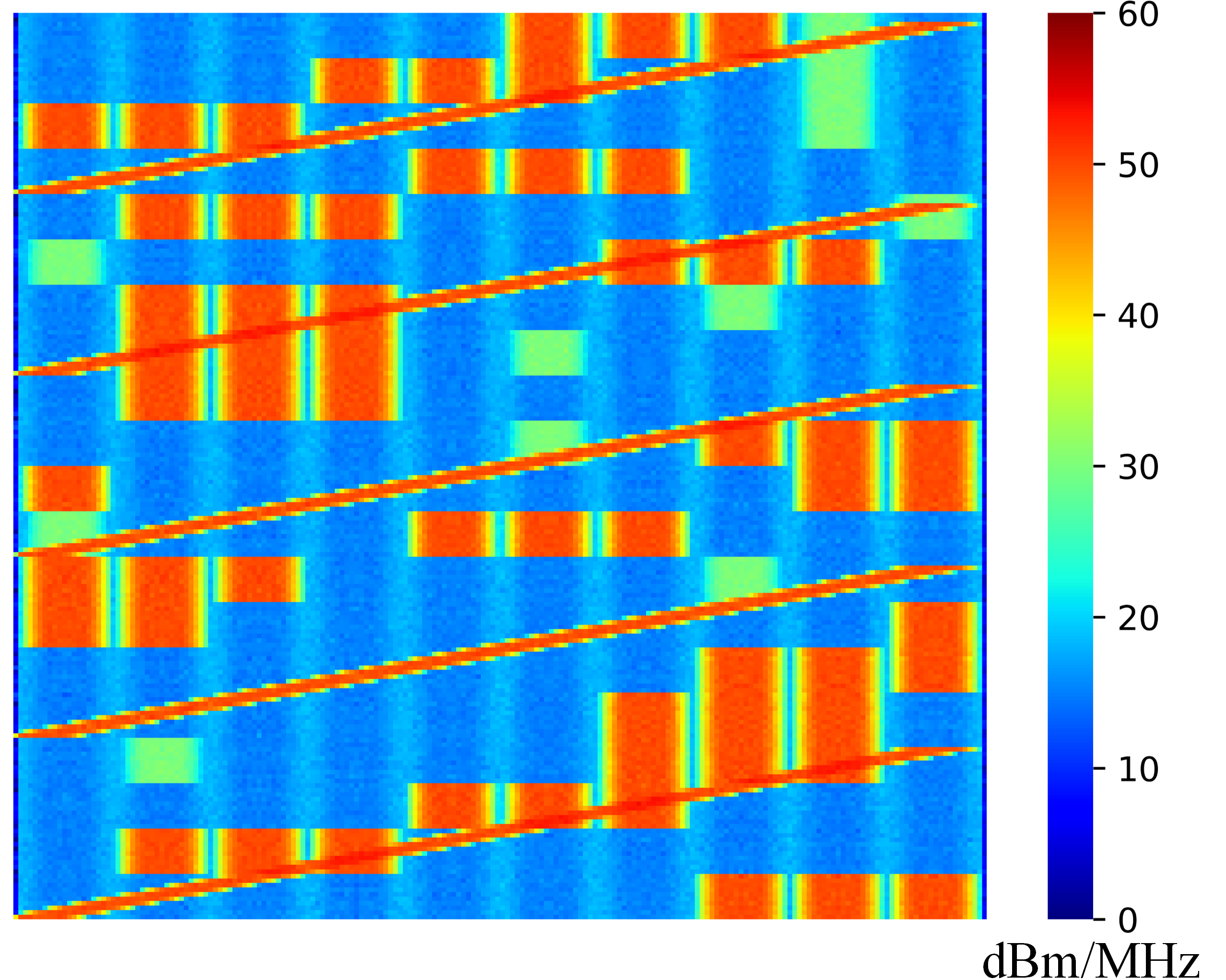}
\caption{A detailed example of the environmental state.}
\label{fig_4}
\end{figure}

For the considered anti-jamming MG, we assume that the actions of both the legitimate user and the intelligent jammer are allowed to change among hops. Since the spectrum waterfall defined in \cite{7423401} contains time, frequency and power domain information, it could be used to represent the complex spectrum state and provide enough information for either anti-jamming or jamming decision-making. Therefore, the environment state of the $n$-th hop could be represented as
\begin{equation}
{\bf{S}}_n = \begin{bmatrix}
{\bf{s}}_{(n-1)N_h-N_T+1} \\
\vdots \\
{\bf{s}}_{(n-1)N_h-1} \\
{\bf{s}}_{(n-1)N_h}
\end{bmatrix}, \tag{11}\label{11}
\end{equation}
where ${\bf{s}}_{k}$ denotes the spectrum vector during the $k$-th time slot, $N_T$ denotes the length of historical data. The state ${\bf{S}}_n \in \mathcal{S}$ is a two-dimensional matrix of size $N_F \times N_T$, with each row corresponding to the channel power measurements during a given time slot. The thermodynamic chart of the matrix ${\bf{S}}_n$ is referred to as the spectrum waterfall \cite{8314744}, which captures both frequency and temporal information, and all possible observed spectrum waterfalls constitute the set of environmental states. Fig. \ref{fig_4} provides a detailed example of the environmental state in the anti-jamming scenario under consideration. The user takes ${\bf{S}}_n$ as the input for anti-jamming decision-making. Similarly, the DRL-based intelligent jammer also takes the observed spectrum waterfall ${\bf{S}}_n^j$ as the input for its jamming decisions. For convenience, ${\bf{S}}_n^j = {\bf{S}}_n$ is used in what follows.

Let $\pi(\cdot)$ and $\mu(\cdot)$ denote the policies of the legitimate user and intelligent jammer, respectively. During the $n$-th hop, the legitimate user executes an anti-jamming action $a_n^u \in \mathcal{F}$ based on the policy $\pi({\bf{S}}_n)$, while the intelligent jammer performs a jamming action $a_n^j \in \mathcal{F}$ according to the policy $\mu({\bf{S}}_n)$. Then, the state ${\bf{S}}_n$ transits to the next state ${\bf{S}}_{n+1}$ with probability $\mathcal{P}({\bf{S}}_{n+1} | {\bf{S}}_n, a_n^u, a_n^j)$, and the environment provides immediate rewards $r_n^u = \mathcal{R}^u({\bf{S}}_n, a_n^u, a_n^j)$ and $r_n^j = \mathcal{R}^j({\bf{S}}_n, a_n^u, a_n^j)$ to the legitimate user and the intelligent jammer, respectively.

At the end of each hop, the receiver responds to the transmitter with a feedback (i.e., ACK/NACK) through the control link. Specifically, if $\beta(k, f_k^u) \geq \beta_\text{th}, \  \forall k = (n-1)N_h+1, \ldots, nN_h-1, nN_h$ holds during the $n$-th hop, the receiver transmits an ACK signal to the transmitter. Otherwise, an NACK signal is transmitted. Motivated by \cite{10308595}, we assume that the legitimate user's agent is located at the receiver. The agent could determine which channel to access and transmit the channel decision message $a_n^u$ to the transmitter at the beginning of the $n$-th hop through the control link. Then it takes the minimum SINR during the $n$-th hop as the immediate reward for action $a_n^u$, namely,
\begin{equation}
r^u_n = \min \left \{ \beta(k, f_u^k) \right \}_{k = (n-1)N_h+1, \ldots, nN_h-1, nN_h}. \tag{12}\label{12}
\end{equation}

Whereas the intelligent jammer has a completely contrasting objective. Although the jammer typically lacks knowledge of the received SINR and demodulation threshold of the legitimate user, the ACK/NACK feedback transmitted from the legitimate receiver to the legitimate transmitter could reflect whether the legitimate signal transmitted during a hop has been successfully received \cite{8558114}. It is assumed that the worst jammer could accurately assess the effectiveness of jamming attacks, and the reward for the intelligent jammer's action $a_n^j$ is determined by the intercepted ACK/NACK signal from the control link during the $n$-th hop, namely,
\begin{align}
r^j_n = \begin{cases}
1, \quad &\text{if the NACK feedback is detected}, \notag \\
-1, \quad &\text{if the ACK feedback is detected}. \tag{13}\label{13}
\end{cases}
\end{align}

In the process of the anti-jamming game, the legitimate user aims to maximize its cumulative reward $R^u({\bf{S}}_n)$, i.e.,
\begin{equation}
R^u({\bf{S}}_1)  = \sum_{n=1}^{\infty} \gamma^n r^u_n. \tag{14}\label{14}
\end{equation}
Meanwhile, the intelligent jammer aims to maximize its cumulative reward $R^j({\bf{S}}_n)$, i.e.,
\begin{equation}
R^j({\bf{S}}_1)  = \sum_{n=1}^{\infty} \gamma^n r^j_n. \tag{15}\label{15}
\end{equation}

In Markov Games, no player’s policy is inherently optimal, as its return is influenced by the actions of other players \cite{9292435}. The best response (BR) and the Nash equilibrium (NE) are commonly used to evaluate the performance of one player against others in MGs. For the aforementioned anti-jamming MG with a legitimate user and an intelligent jammer, the BR and the NE are defined as follows. 
\begin{myDef}(BR in the anti-jamming MG)
\label{D.1}
Given the policy of the intelligent jammer $\mu$, the policy $\pi^b$ of the user is defined as the BR policy if there exists no alternative policy that could yield a higher cumulative reward, formally expressed as
\begin{equation}
R^u({\bf{S}}_1; \pi^b, \mu) \geq R^u({\bf{S}}_1; \pi, \mu), \quad \forall \pi. \tag{16}\label{16}
\end{equation}
Conversely, when the user employs the policy $\pi$, the BR policy of the intelligent jammer, denoted as $\mu^b$, must satisfy
\begin{equation}
R^j({\bf{S}}_1; \pi, \mu^b) \geq R^j({\bf{S}}_1; \pi, \mu), \quad \forall \mu. \tag{17}\label{17}
\end{equation}
\end{myDef}

\begin{myDef}(NE in the anti-jamming MG)
\label{D.2}
The NE is defined as a pair of policies $(\pi^*, \mu^*)$, where $\pi^*$ and $\mu^*$ are mutual best responses. Formally, this could be expressed as
\begin{align}
R^u({\bf{S}}_1; \pi, \mu^*) &\leq R^u({\bf{S}}_1; \pi^*, \mu^*) \leq R^u({\bf{S}}_1; \pi^*, \mu), \quad \forall \pi, \mu; \notag \\
R^j({\bf{S}}_1; \pi^*, \mu) &\leq R^j({\bf{S}}_1; \pi^*, \mu^*) \leq R^j({\bf{S}}_1; \pi, \mu^*), \quad \forall \pi, \mu. \tag{18}\label{18}
\end{align}
\end{myDef}

According to \cite{nash1951non}, the NE of the proposed MG always exists and is equivalent to the minimax solution of the game, i.e.,
\begin{align}
R^u({\bf{S}}_1; \pi^*, \mu^*) = \max_\pi \min_\mu R^u({\bf{S}}_1; \pi, \mu) = \min_\mu \max_\pi R^u({\bf{S}}_1; \pi, \mu); \notag \\
R^j({\bf{S}}_1; \pi^*, \mu^*) = \max_\mu \min_\pi R^j({\bf{S}}_1; \pi, \mu) = \min_\pi \max_\mu R^j({\bf{S}}_1; \pi, \mu). \tag{19}\label{19}
\end{align}
The NE delineates the maximum return that a legitimate user could achieve when facing a formidable opponent. It is particularly meaningful when the jammer is capable of adapting its policy in response to the user's actions. When the legitimate user adopts the policy $\pi^*$, its return is guaranteed to be at least equal to the Nash equilibrium. While, if the legitimate user switches to an alternative anti-jamming policy, its return might fall below $R^u({\bf{S}}_n; \pi^*, \mu^*)$.

Minimax Q learning \cite{5738229} could be employed to determine the NE for MGs without any prior knowledge of the environment dynamics. Additionally, the legitimate user could achieve a beyond NE performance by employing an opponent modeling based DQN \cite{9939159}. However, these methods assume that the legitimate user could access the actions selected by the intelligent jammer at the beginning of each hop, which is generally impractical. Meanwhile, these DRL-based methods necessitate extensive training episodes in non-stationary environments, where the strategies of both the legitimate user and the intelligent jammer could continuously evolve over time. To address these issues, we propose a fast adaptive anti-jamming channel access approach to find a policy beyond NE with fewer training episodes in the following sections.

\section{Coarse-Grained Spectrum Prediction}
Since the legitimate user could not directly observe the jammer's actions in practical anti-jamming scenarios, inferring the intelligent jammer's policy using existing opponent modeling approaches becomes challenging. Although the actions of the intelligent jammer are not accessible in practice, the spectra observed during the current hop could still reflect the behavior of both the legitimate user and the jammers. Therefore, it is possible to employ a convolutional neural network (CNN) model for predicting the spectrum state of the current hop, thereby enhancing the learning capability of the legitimate user's agent. Since the resolution of the spectrum vector defined in eq. (\ref{5}) is too fine and thus inefficient for spectrum prediction, we employ the coarse-grained (CG) spectrum with only $M$ samples for each hop.

\begin{figure*}[!htb]
\centering
\includegraphics[width=5.5in]{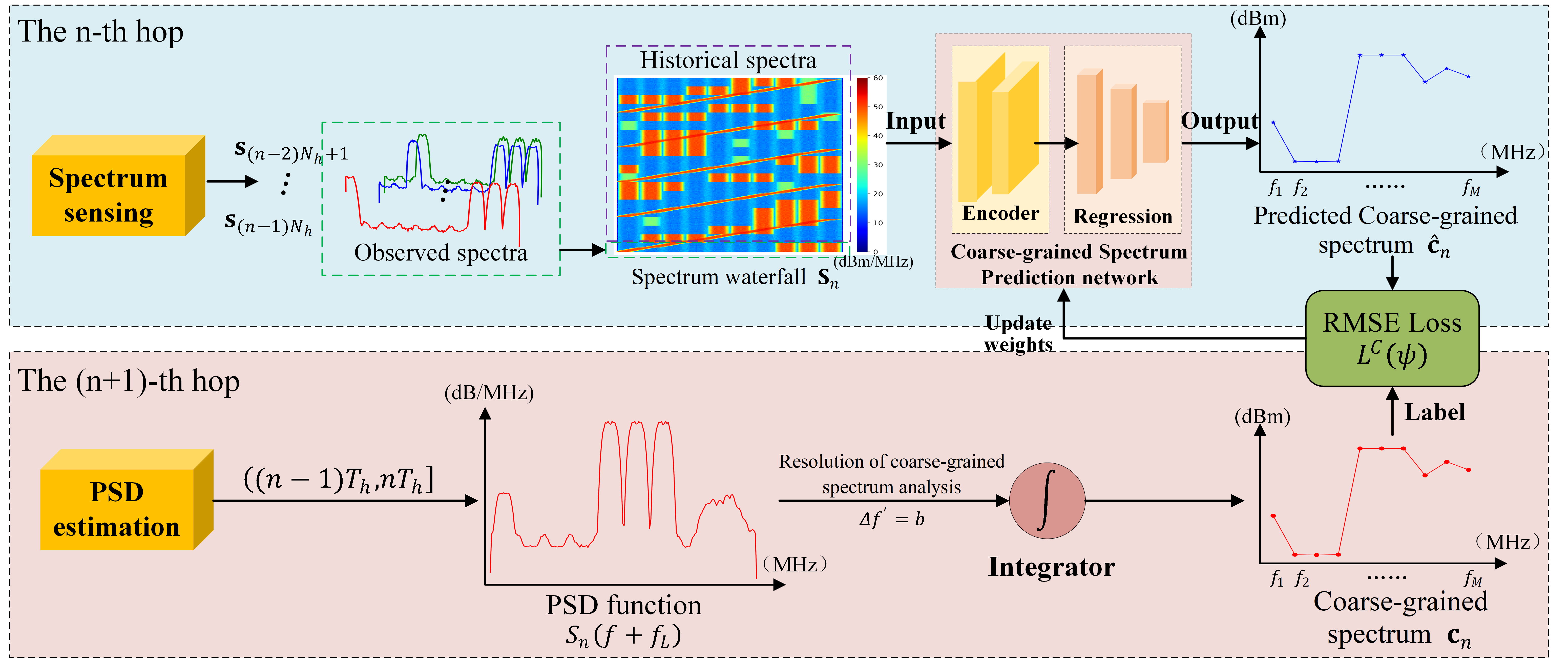}
\caption{Overall structure of the proposed coarse-grained spectrum prediction.}
\label{fig_5}
\end{figure*}

Intuitively, we formulate the coarse-grained spectrum prediction as a regression problem that could be addressed using a supervised learning model $F(\cdot; \psi)$, i.e.,
\begin{equation}
\hat{\bf{c}}_{n} = F \left ({\bf{S}}_n; \psi \right ). \tag{20}\label{20}
\end{equation}
where the input ${\bf{S}}_n$ is the observed spectrum waterfall at the beginning of the current hop, and the output $\hat{\bf{c}}_{n}$ is the predicted coarse-grained spectrum during the current hop. The supervised learning model $F(\cdot; \psi)$ could be implemented by a CNN, where $\psi$ denotes the collection of parameters for the CNN model. As illustrated in Fig. \ref{fig_5}, the proposed CNN model is composed of two convolutional (Conv) layers and three fully connected (FC) layers. The Conv layers process the input spectrum waterfall, while the FC layers integrate the processed information. It is worth noting that the architecture of the network depicted is just an example and could be tailored to suit the specific scenario.

During the training procedure, the legitimate user collects the spectrum waterfall samples and the ground-truth labels (i.e., the corresponding coarse-grained spectra) from the real-time interaction between the legitimate user and the wireless adversarial environment. The collected sample-label pair $({\bf{S}}_n, {\bf{c}}_{n})$ is then stored in the memory $\mathcal{D}_C$. However, since the user's anti-jamming policy and the jammer's policy are continuously updated through their respective agents during the interactions, the regression model trained on a fixed training dataset may not perform well in a dynamically changing environment. This is because legitimate users with varying anti-jamming policies may take entirely different actions when observing the same spectrum waterfall, and the same holds for an intelligent jammer. To address the non-stationarity characteristic of the environment and enable the model to adapt to recent changes in the behavior of both the legitimate user and the jammers, the spectrum prediction model is not trained on the entire accumulated dataset. Instead, a dynamically refreshed memory is employed, which follows the first-in-first-out (FIFO) principle and stores the most recent $|\mathcal{D}_C|$ sample-label pairs for training the regression model $F(\cdot; \psi)$.

In each training epoch, the memory $\mathcal{D}_C$ is fed to the regression model $F(\cdot; \psi)$ for training, where one epoch corresponds to a complete pass through the entire training dataset. The parameters $\psi$ of the regression model could be updated by stochastic gradient descent (SGD), i.e.,
\begin{equation}
\psi_{n+1} = \psi_n - \alpha_C \nabla_{\psi_n}L^C(\psi), \tag{21}\label{21}
\end{equation}
where $\alpha_C$ is the learning rate and $L^C(\psi)$ is the loss function. To minimize the expected regression error, the standard root mean squared error (RMSE) loss is adopted as
\begin{equation}
L^C(\psi) = \mathbb{E}_{\mathcal{B}_C \subseteq \mathcal{D}_C} \left [ \left \| \hat{\bf{c}}_i - {\bf{c}}_i \right \| \right ] = \mathbb{E}_{\mathcal{B}_C \subseteq \mathcal{D}_C} \left [ \sqrt{\sum_{m=1}^M \left ( \hat{c}_i^m - c_i^m \right )^2} \right ], \tag{22}\label{22}
\end{equation}
where
\begin{equation}
\hat{\bf{c}}_i = F \left ({\bf{S}}_{i}; \psi \right ) = \left [ \hat{c}_i^1, \hat{c}_i^2, \ldots, \hat{c}_i^M \right ], \tag{23}\label{23}
\end{equation}
denotes the output of the CG spectrum prediction model, and 
\begin{equation}
{\bf{c}}_i = \left [ c_i^1, c_i^2, \ldots, c_i^M \right ], \tag{24}\label{24}
\end{equation}
denotes the ground-truth label of the sample ${\bf{S}}_{i}$. It is noted that $({\bf{S}}_{i}, {\bf{c}}_i) \in \mathcal{D}_C$ is a sample in $\mathcal{D}_C$, where ${\bf{S}}_{i}$ is the spectrum waterfall composed of several spectrum vectors from previous hops, and ${\bf{c}}_i$ is the coarse-grained spectrum of the current hop. This enables the network to predict the future coarse-grained spectrum based on the currently observed spectrum waterfall. In order to enable the proposed coarse-grained spectrum prediction model to adapt to the non-stationary nature of the environment, we do not use all the accumulated data, as is typically done in conventional supervised learning, for model training. Instead, at each training iteration, only the most recent $| \mathcal{D}_C |$ samples are employed to update the parameters of the coarse-grained spectrum prediction model. This strategy ensures that the prediction model remains sensitive to the latest changes in the environment, including the evolving transmission behavior of legitimate users and adversarial jamming strategies. By focusing on recent samples, the model could better track the dynamic spectrum patterns and maintain high prediction accuracy in non-stationary scenarios.

\section{Fast Adaptive Anti-jamming Channel Access Approach}
As previously discussed, the environmental dynamics are influenced by multiple agents. To identify the dynamic BR to the time-varying jamming policy, the legitimate user's agent needs to extract knowledge from the environment and exploit the opponent's behavior as soon as possible in a multi-agent system. To this end, we propose a fast adaptive anti-jamming channel access approach with joint DQN and coarse-grained spectrum prediction, where the coarse-grained spectrum prediction is adopted as an auxiliary task for learning the jammer's policy feature and environment dynamics.

\begin{figure*}[!htb]
\centering
\includegraphics[width=5.5in]{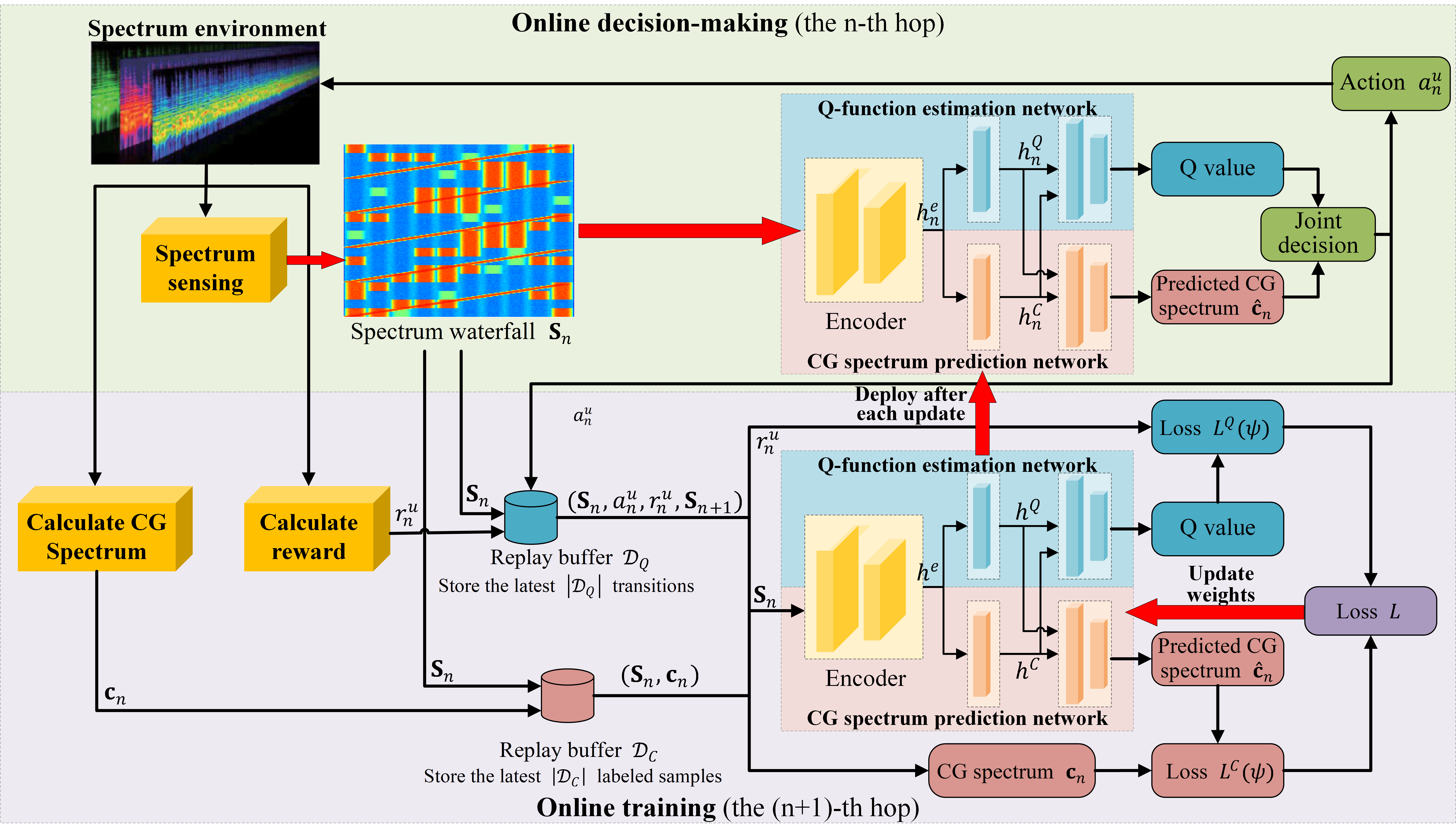}
\caption{Overall structure of the fast adaptive anti-jamming approach with joint DQN and coarse-grained spectrum prediction.}
\label{fig_6}
\end{figure*}

As shown in Fig. \ref{fig_6}, we construct a Q-function estimation network and a CG spectrum prediction network for determining the legitimate transmitter's channel access action at the beginning of each hop. Specifically, the Q-function estimation network consists of a feature extraction module and an inference module. Meanwhile, the coarse-grained spectrum prediction network is composed of a feature extraction module and a regression forecasting module. It is noted that both the Q-function estimation network and the CG spectrum prediction network take the spectrum waterfall ${\bf{S}}_n$ as the input, then output the estimated Q-value and the predicted CG spectrum, respectively. The legitimate user's anti-jamming action is determined based on the Q-value $Q({\bf{S}}_n, a_n^u), \forall a_n^u \in \mathcal{F}$ and the predicted CG spectrum $\hat{\bf{c}}_n$, jointly. The architecture and the training procedure of the proposed fast adaptive anti-jamming channel access approach are discussed in what follows.

\subsection{Q-function Estimation}
Since the jammer's actions are unknown to the legitimate user in practical scenarios, we simplify the proposed anti-jamming MG $\mathcal{G} = \left \{ \mathcal{S}, \mathcal{A}^u, \mathcal{A}^j, \mathcal{P}, \mathcal{R}^u, \mathcal{R}^j, \gamma \right \}$ to an anti-jamming MDP, which could be addressed using RL-based methods. The anti-jamming MDP could be formulated by a five-tuple $\left \{ \mathcal{S}, \mathcal{A}^u, \mathcal{P}, \mathcal{R}^u, \gamma \right \}$, where $\mathcal{S}$ is the set of environment state, $\mathcal{A}^u$ is the legitimate user's action set, $\mathcal{P}$ denotes the transition function, $\mathcal{R}^u$ denotes the legitimate user's reward function, $\gamma \in (0, 1]$ denotes the discount factor.

The legitimate user aims to identify a policy $\pi$ that maximizes the expected cumulative reward, and the Q-function of a given policy $\pi$ is defined as the expected cumulative reward starting from a state-action pair $({\bf{S}}, a^u)$, namely,
\begin{equation}
Q_\pi({\bf{S}}, a^u) = \mathbb{E}_\pi \left [ \sum_{n=1}^\infty \gamma^n r_n^u | {\bf{S}}_1 = {\bf{S}}, a_1^u = a^u \right ]. \tag{25}\label{25}
\end{equation}
The optimal Q-function $Q^*({\bf{S}}, a^u)$ could be calculated through the Bellman optimality equation \cite{sutton1998introduction}, i.e.,
\begin{align}
Q^*({\bf{S}}, a^u) = \sum_{{\bf{S}}^\prime} \mathcal{P}({\bf{S}}^\prime | {\bf{S}}, a^u) \left [ r^u + \gamma \max_{{a^u}^\prime} Q^*({\bf{S}}^\prime, {a^u}^\prime) \right ], \tag{26}\label{26}
\end{align}
where ${a^u}^\prime$ is the legitimate user's action to be selected in state ${\bf{S}}^\prime$. The agent could iteratively collect transition $({\bf{S}}_n, a_n^u, r_n^u, {\bf{S}}_{n+1})$ from the environment and store it in memory $\mathcal{D}_Q$. Then, the objective of the user could be simplified to the determination of the optimal value of Q-function for all state-action pairs. Therefore, the legitimate user's optimal strategy could be formulated as
\begin{equation}
\pi^*({\bf{S}}) = \mathop{\arg\max}\limits_{\pi(a^u | {\bf{S}})} Q^*({\bf{S}}, a^u). \tag{27}\label{27}
\end{equation}

Since the environment state evolves dynamically in accordance with the policy $\pi$, and the state-action space is extensive in the proposed anti-jamming MDP, a model-free RL algorithm based on deep neural networks, i.e., DQN \cite{mnih2015human}, is employed for estimating the Q-function over the high-dimensional and complex state space. Here, we utilize a CNN-based model as the Q network to approximate the Q-function for each state-action pair $({\bf{S}}_n, a_n^u)$, i.e.,
\begin{equation}
Q({\bf{S}}_n, a_n^u; \theta) = \mathbb{E}\left [r_n^u + \gamma \max_{a_{n+1}^u} Q\left ({\bf{S}}_{n+1}, a_{n+1}^u; \theta | {\bf{S}}_n, a_n^u \right ) \right ], \tag{28}\label{28}
\end{equation}
where $\theta$ denotes the parameters of the Q network and could be optimized by minimizing the loss function $L^Q(\theta)$ using gradient descent. $L^Q(\theta)$ is calculated by a random batch of transitions (i.e., $\mathcal{B}_Q \subseteq \mathcal{D}_Q$) and could be expressed as
\begin{equation}
L^Q(\theta) = \mathbb{E}_{\mathcal{B}_Q \subseteq \mathcal{D}_Q} \left [ \left ( \eta_n - Q({\bf{S}}_n, a_n^u; \theta) \right)^2 \right ], \tag{29}\label{29}
\end{equation}
where 
\begin{equation}
\eta_n = r_n^u + \gamma \max_{a_{n+1}^u} Q\left ({\bf{S}}_{n+1}, a_{n+1}^u; \theta^- \right ), \tag{30}\label{30}
\end{equation}
is the target Q value of each transition in $\mathcal{D}_Q$. It is noted that the target network mirrors the Q network, with the exception that its parameters $\theta^-$ could be updated by the Q network at a predetermined interval (i.e., $\theta^-$ is updated every $N_u$ hops).

\subsection{Joint Inference and Decision-Making}
To identify the anti-jamming BR policy for the legitimate user, we propose a joint inference and decision-making method aimed at enhancing the state feature representations of the legitimate user's agent in the proposed anti-jamming scenario. The legitimate user could learn the jammer’s policy features through an auxiliary task (i.e., CG spectrum prediction) and the jammer’s policy features could serve as a hidden representation for inferring the intelligent jammer's behavior. Given that the hidden representation encodes the spatial-temporal features of the jammer’s policy $\mu$ and could be learned from the user's observations over a series of consecutive hops, incorporating the hidden representation into the DQN model could assist in identifying a superior Q-function compared to traditional DRL methods. Similarly, the hidden features extracted by the DQN model could also enhance the learning capability of the coarse-grained spectrum prediction model.

\begin{figure}[!t]
\centering
\includegraphics[width=2.3in]{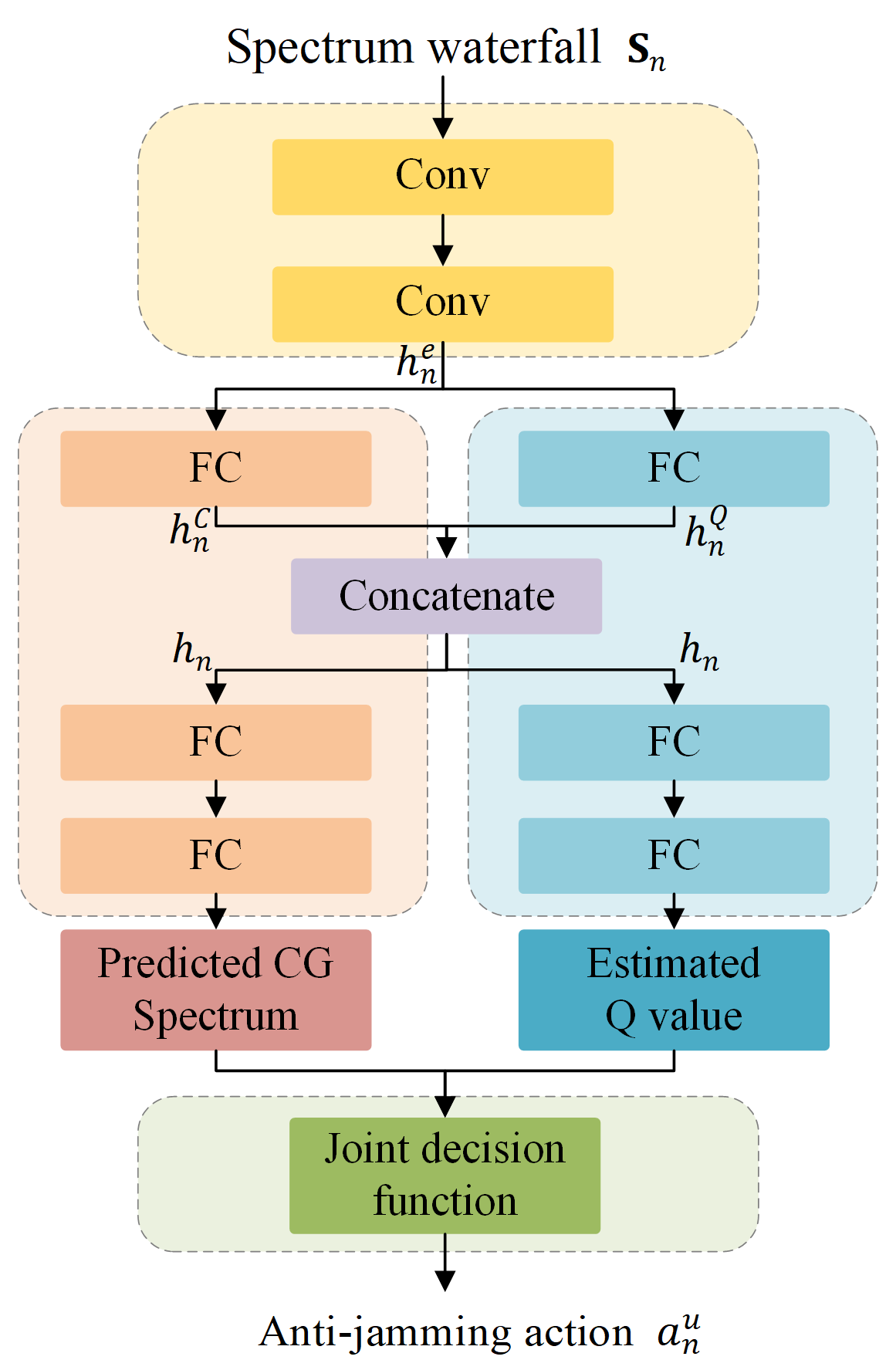}
\caption{The network architecture of the joint inference and decision-making model.}
\label{fig_7}
\end{figure}

We design a dual-branch CNN architecture to jointly perform coarse-grained spectrum prediction and Q-value estimation, as illustrated in Fig. \ref{fig_7}. The input spectrum waterfall ${\bf{S}}_n$ is initially processed by two shared convolutional layers to extract time-frequency features, resulting in a latent representation, denoted as $h_n^e$. This representation is then fed into two parallel branches: one for spectrum prediction and the other for decision evaluation. In the prediction branch, $h_n^e$ is processed by an FC layer to produce $h_n^C$. In the evaluation branch, $h_n^e$ is similarly transformed into $h_n^Q$ via an FC layer. Then the two representations $h_n^Q$ and $h_n^C$ are concatenated to form a joint feature representation $h_n$, which captures both prediction and evaluation cues. This shared representation is then separately fed into task-specific FC layers in the two branches to generate the coarse-grained spectrum prediction and estimate the Q-value, respectively. Such a design facilitates effective feature sharing while maintaining task-specific outputs.

It is important to note that the feature extraction module is shared between the coarse-grained spectrum prediction module and the Q-function estimation module. Subsequently, these two modules take $h_n^e$ as their common input, respectively. Both the coarse-grained spectrum prediction module and the Q-function estimation module consist of several FC layers. The coarse-grained spectrum prediction module is trained to infer the CG spectrum during the current hop, while the Q-function estimation module is trained to approximate the optimal Q-function. The two modules operate independently to extract two distinct types of features (i.e., $h_n^C$ and $h_n^Q$) from $h_n^e$ using an FC layer, and these features are concatenated to construct the hidden representation $h_n = \left ( h_n^Q; h_n^C \right )$. After that, $h_n$ is processed by two FC layers to generate the predicted coarse-grained spectrum during the current hop (i.e., $\hat{\bf{c}}_n$) within the coarse-grained spectrum prediction module and the estimated Q-values (i.e., $Q({\bf{S}}_n, a^u), \ \forall a^u \in \mathcal{A}^u$) within the Q-function estimation module, respectively. Ultimately, the legitimate user's anti-jamming channel access action is determined by
\begin{equation}
a_n^u = \mathop{\arg\max}\limits_{a^u} F_a\left ( \hat{\bf{c}}_n, Q({\bf{S}}_n, a^u)  \right ), \tag{31}\label{31}
\end{equation}
where $F_a(\cdot)$ represents a joint decision function involving the predicted coarse-grained spectrum $\hat{\bf{c}}_n$ and Q-values $Q({\bf{S}}_n, a^u), \ \forall a^u \in \mathcal{A}^u$. Furthermore, one of the possible joint decision functions could take the form of 
\begin{equation}
F_a\left ( \hat{\bf{c}}_n, Q({\bf{S}}_n, a^u)  \right ) =  \frac{Q({\bf{S}}_n, a^u)}{10^{\hat{\bf{c}}_n[a^u]/10}}, \tag{32}\label{32}
\end{equation}
where $Q({\bf{S}}_n, a^u)$ denotes the predicted Q-value of channel $a^u$ under the current state ${\bf{S}}_n$, and $\hat{\bf{c}}_n[a^u]$ represents the predicted coarse-grained spectrum sample (in dBm) for channel $a^u$. This function is intended to prioritize channels that simultaneously offer high expected long-term reward and low predicted spectrum sample value. The denominator converts the spectrum sample value from dBm to mW, thereby aligning it with the magnitude of the Q-value in the numerator and ensuring that both components contribute comparably to the decision-making process. It is noted that this decision-making function is merely one of the viable functions, which could be further refined to suit the specific anti-jamming scenario.

It is noted that, in the proposed joint decision mechanism, the coarse-grained spectrum prediction results are not used as a standalone decision but instead serve as an input to the joint decision module. During the early training stages, these predictions help improve the quality of action selection over random exploration. As a result, the agent is more likely to obtain positive reward samples early on, which in turn accelerates the rate of convergence of the Q-network and enhances the anti-jamming performance. This design improves the exploration efficiency of the DRL framework without compromising its learning capability.

The training procedure for the proposed joint inference and decision-making model is detailed in Algorithm \ref{alg1}, where $L^C_\text{th}$ serves as the threshold to balance exploration and exploitation for the anti-jamming agent. We incorporate two distinct loss terms (i.e., $L^C(\psi)$ defined in eq. (\ref{22}) and $L^Q(\theta)$ defined in eq. (\ref{29})) to construct an aggregated loss function $L(\theta, \psi)$ for training the proposed anti-jamming model, jointly. The aggregated loss $L(\theta, \psi)$ is defined as
\begin{equation}
L(\theta, \psi) = \lambda L^Q(\theta) + L^C(\psi), \tag{33}\label{33}
\end{equation}
where $L^C(\psi)$ is the regression forecasting loss for coarse-grained spectrum prediction, $L^Q(\theta)$ is the standard DQN loss function, $\lambda = \frac{1}{\sqrt{L^C(\psi)}}$ serves as the adaptive scale factor of $L^Q(\theta)$, which could adaptively adjust the scaling of $L^Q(\theta)$ during different phases of the training process. Unlike conventional fixed-weight strategies, our framework adaptively adjusts $\lambda$ during the training process. This allows the model to prioritize prediction accuracy during early training and gradually emphasize value estimation as learning progresses. Such dynamic re-weighting enhances training stability and fosters better synergy between the two learning objectives.

\begin{algorithm}[!t]
\caption{The Fast Adaptive Anti-Jamming Approach with Joint DQN and CG Spectrum Prediction.}
\begin{algorithmic}
\STATE Initialize the memory $\mathcal{D}_Q = \emptyset$ and $\mathcal{D}_C = \emptyset$.
\STATE Initialize the CG spectrum prediction network with random weights $\psi$ and $L^C(\psi) = 100$.
\STATE Initialize the Q network with random weights $\theta$.
\STATE Initialize the target Q network with weights $\theta^- = \theta$.
\STATE Observe the initial state ${\bf{S}}_1$.
\FOR{$n = 1, 2, \ldots, \infty$}
\IF{$L^C(\psi) > 10$}
\STATE The legitimate user chooses a random anti-jamming action $a_n^u \sim \mathcal{A}^u$.
\ELSE 
\STATE The legitimate user chooses anti-jamming action $a_n^u$ according to (\ref{31}).
\ENDIF
\STATE The intelligent jammer selects jamming action $a_n^j \sim \mu_n({\bf{S}}_n)$.
\STATE The user switches to channel $a_n^u$ for legitimate transmission, while the intelligent jammer transmits jamming signals according to the jamming action $a_n^j$.
\STATE Calculate the immediate reward $r_n^u$ and the CG spectrum ${\bf{c}}_n$ for the current hop.
\STATE Observe the next state ${\bf{S}}_{n+1}$.
\STATE Update $\mathcal{D}_Q$ with transition $({\bf{S}}_n, a_n^u , r_n^u, {\bf{S}}_{n+1})$ following the FIFO principle.
\STATE Update $\mathcal{D}_C$ with sample-label pair $({\bf{S}}_n, {\bf{c}}_n)$ following the FIFO principle.
\IF{$Sizeof(\mathcal{D}_Q) > | \mathcal{B}_Q |$ and $Sizeof(\mathcal{D}_C) > | \mathcal{B}_C |$}
\STATE Sample a minibatch $\mathcal{B}_Q$ randomly from $\mathcal{D}_Q$.
\STATE Sample a minibatch $\mathcal{B}_C$ randomly from $\mathcal{D}_C$.
\STATE Calculate aggregated loss $L(\theta, \psi)$ via (\ref{33}).
\STATE Update Q network with $\theta = \theta - \alpha_Q \nabla_\theta L(\theta, \psi)$.
\STATE Update CG spectrum prediction network with $\psi = \psi - \alpha_C \nabla_\psi L(\theta, \psi)$.
\STATE Update $\theta^- = \theta$ for every $N_u$ steps.
\ENDIF
\ENDFOR
\end{algorithmic}
\label{alg1}
\end{algorithm}

\section{Experiments}
\subsection{Simulation Setup}
Simulations are presented to validate the superiority of the proposed anti-jamming approach, with parameter settings referenced from \cite{8999433, 8314744}. In experiments, the frequency band from 0MHz to 20MHz (i.e., the total bandwidth $B = 20$MHz) is divided into $M=10$ non-overlapping channels, with the bandwidth of each channel being $b=2$MHz. The path loss exponent is set to $\alpha_d = 1$, and the instantaneous fading coefficient $\xi_k$ follows an exponential distribution with a mean of 1 \cite{6542754}. The legitimate receiver and the intelligent jammer could perform spectrum sensing actions every $\Delta t=1$ms with $\Delta f = 100$kHz, retaining the spectrum data for $T = 200$ms. The number of samples in each spectrum vector is $N_F = B / \Delta f = 200$ and the length of historical data is $N_T = T/ \Delta t = 200$. Therefore, the size of the spectrum waterfall ${\bf{S}}_n$ is $N_F \times N_T = 200 \times 200$. The fixed-mode jammer could launch traditional jamming attacks with a predetermined jamming mode, while the intelligent jammer could attack $N_I = 3$ consecutive channels in each time slot, with an update step 10 times that of the legitimate user \cite{9500773, 9939159}. Additionally, the hyper-parameters of the intelligent jammer's agent are set according to \cite{9939159}. The bandwidth of each jamming tone is 2MHz, and its power is set to 50dBm.

The legitimate signal has a bandwidth of 2MHz, and its working frequency is allowed to hop every $T_h = 10$ms. The transmission power of the legitimate transmitter is 30dBm, and the legitimate receiver's demodulation threshold is set to $\beta_\text{th}=0$dB. The legitimate signal is shaped with a root-raised cosine pulse shaping filter with a roll-off factor of 0.5. The hyper-parameters in Algorithm 1 are set as follows: learning rate is $\alpha_Q = \alpha_C = 1 \times 10^{-4}$, minibatch size is $|\mathcal{B}_Q| = 64$, size of memory $\mathcal{D}_Q$ is $|\mathcal{D}_Q| = 1000$, size of memory $\mathcal{D}_C$ is $|\mathcal{D}_C| = 64$, update frequency of target Q network is $N_u = 1000$, discount factor is $\gamma=0.1$, loss threshold is $L^C_\text{th}=10$. The Q-function estimation network and the coarse-grained spectrum prediction network share the same CNN architecture, which is detailed in Table \ref{tab1}. During the training procedure, one episode consists of 100 hops, with the parameters of both the coarse-grained spectrum prediction network and the Q-function estimation network updated at the end of each hop.

\begin{table*}[!htb]
\caption{Architecture of the CNN}
\label{tab1}
\centering
\begin{tabular}{|c|c|c|c|c|}
\hline
Layer & Input   & Output  & Parameters                                                               & Activation \\ \hline
Conv1 & $200\times 200$ & $16\times 100\times 100$ & \begin{tabular}[c]{@{}c@{}c@{}}Kernel:8; Stride:2; Filter:16\end{tabular}  & ReLU       \\ \hline
Conv2 & $16\times 100\times 100$ & $32\times 50\times 50$ & \begin{tabular}[c]{@{}c@{}c@{}}Kernel:4; Stride:2; Filter:32\end{tabular} & ReLU       \\ \hline
FC1   & $32\times 50\times 50$ & $512$     & -                                                                        & ReLU       \\ \hline
FC2   & $512 \times 2$     & $256$       & -                                                                        & ReLU          \\ \hline
FC3   & $256$     & $10$       & -                                                                        & -          \\ \hline
\end{tabular}
\end{table*}

\subsection{Results}
\begin{figure*}[!htb]
\centering 
\subfloat[Sweeping jamming \cite{9733393}.]{\includegraphics[width=2in]{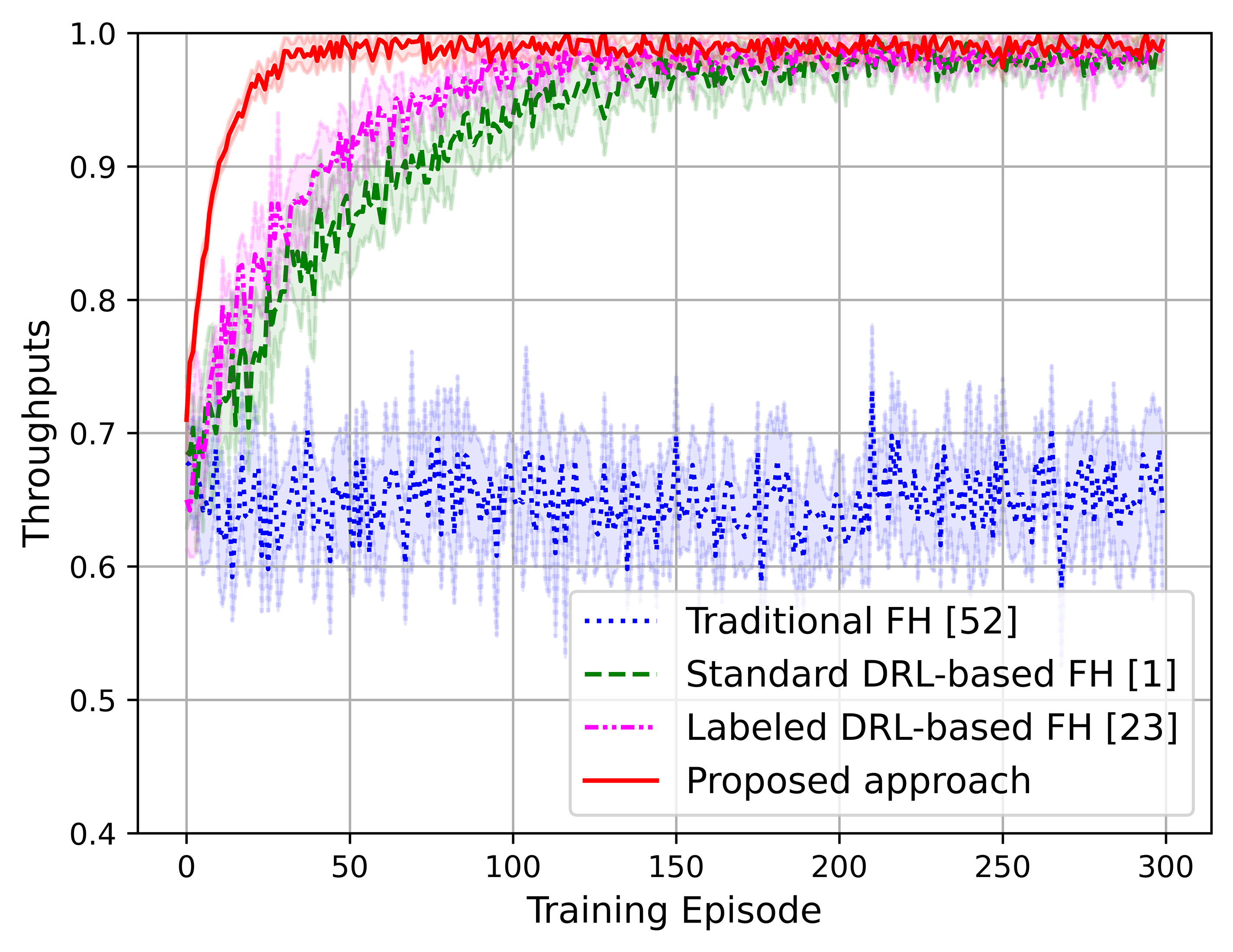}%
\label{8a}}
\hfil
\subfloat[Comb jamming \cite{20075110977801}.]{\includegraphics[width=2in]{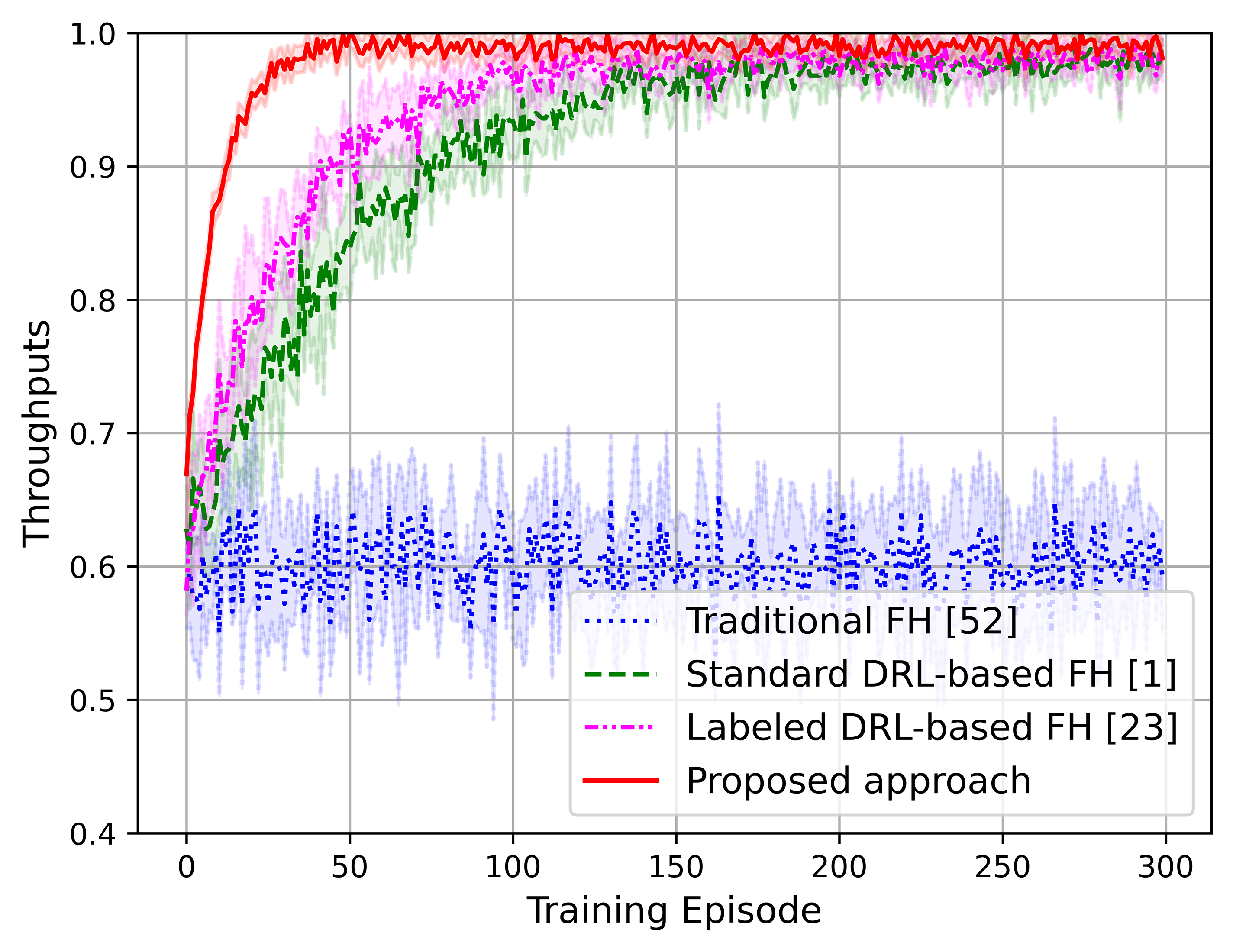}%
\label{8b}}
\hfil
\subfloat[Switch comb jamming \cite{9105045}.]{\includegraphics[width=2in]{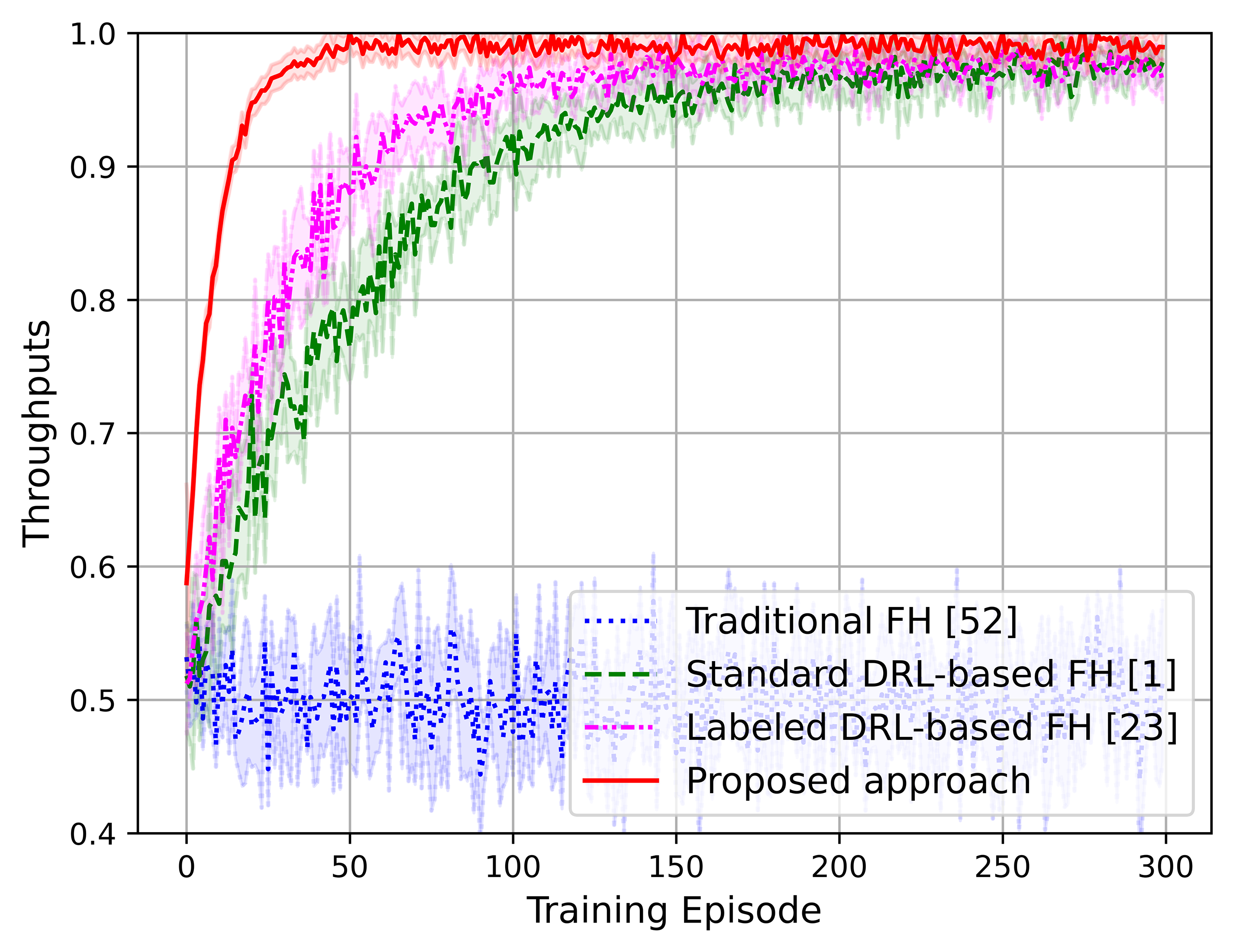}%
\label{8c}}
\hfil
\subfloat[Dynamic jamming \cite{9105045}.]{\includegraphics[width=2in]{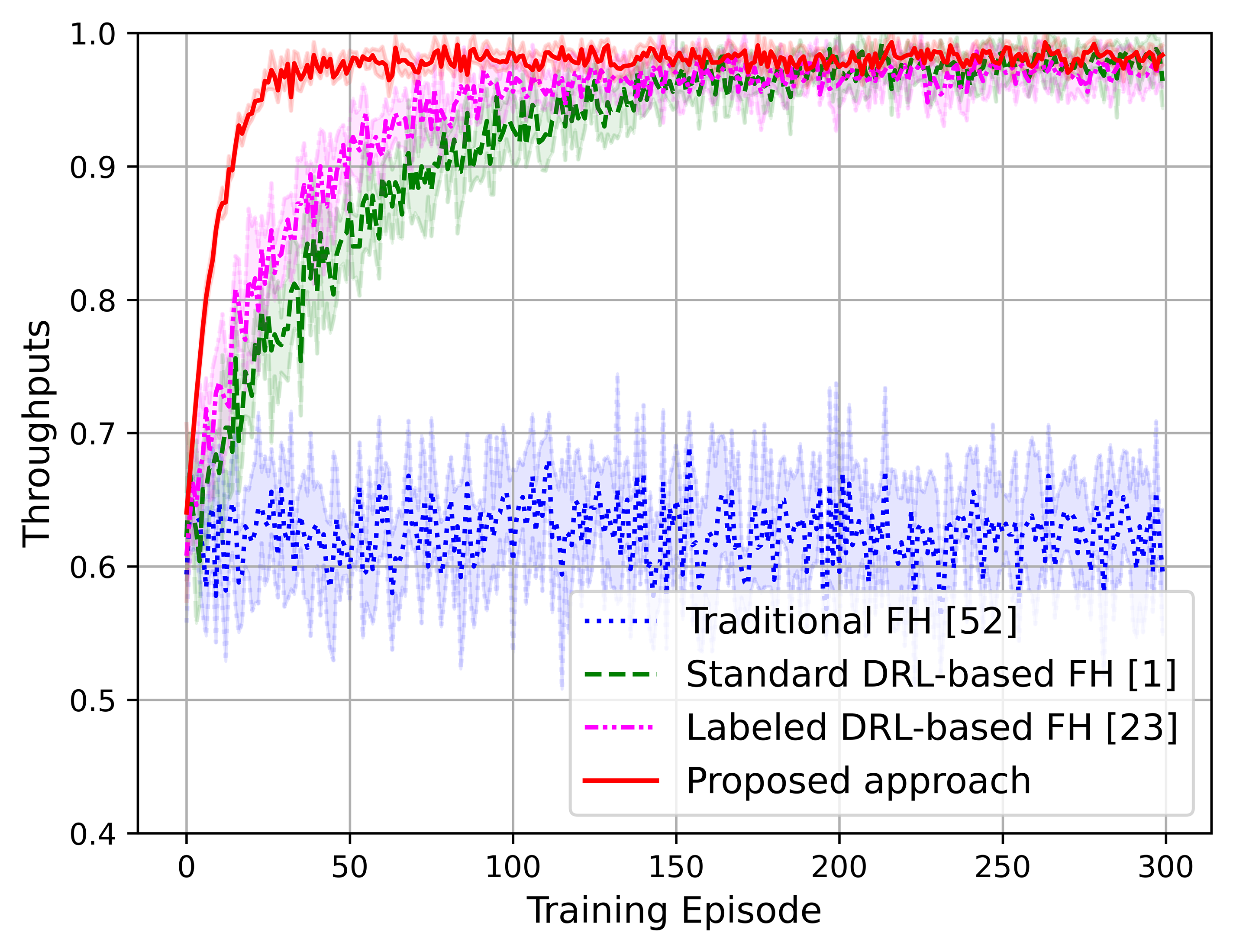}%
\label{8d}}
\hfil
\subfloat[Partial-band jamming \cite{5473884}.]{\includegraphics[width=2in]{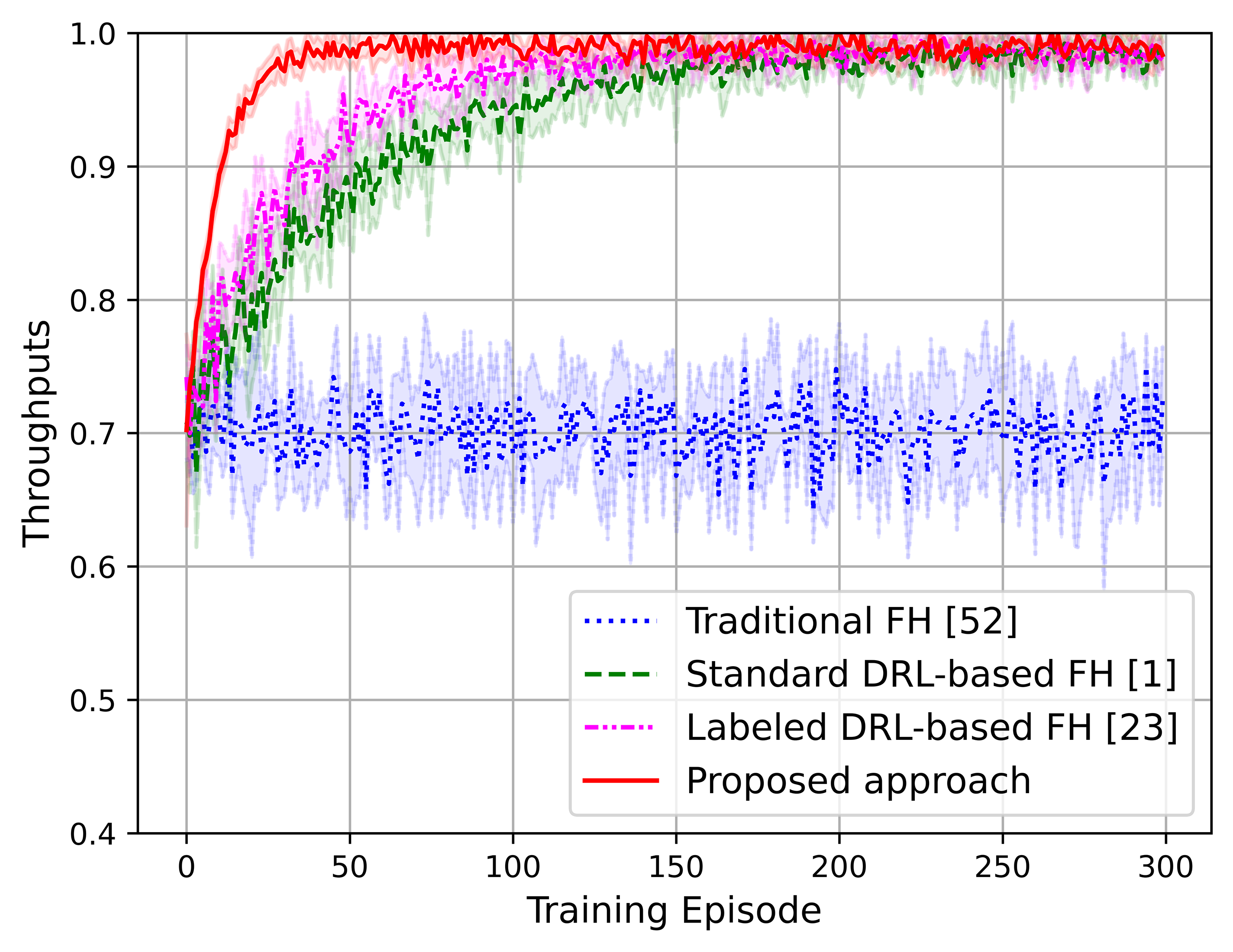}%
\label{8e}}
\hfil
\subfloat[Follower jamming \cite{9721854}.]{\includegraphics[width=2in]{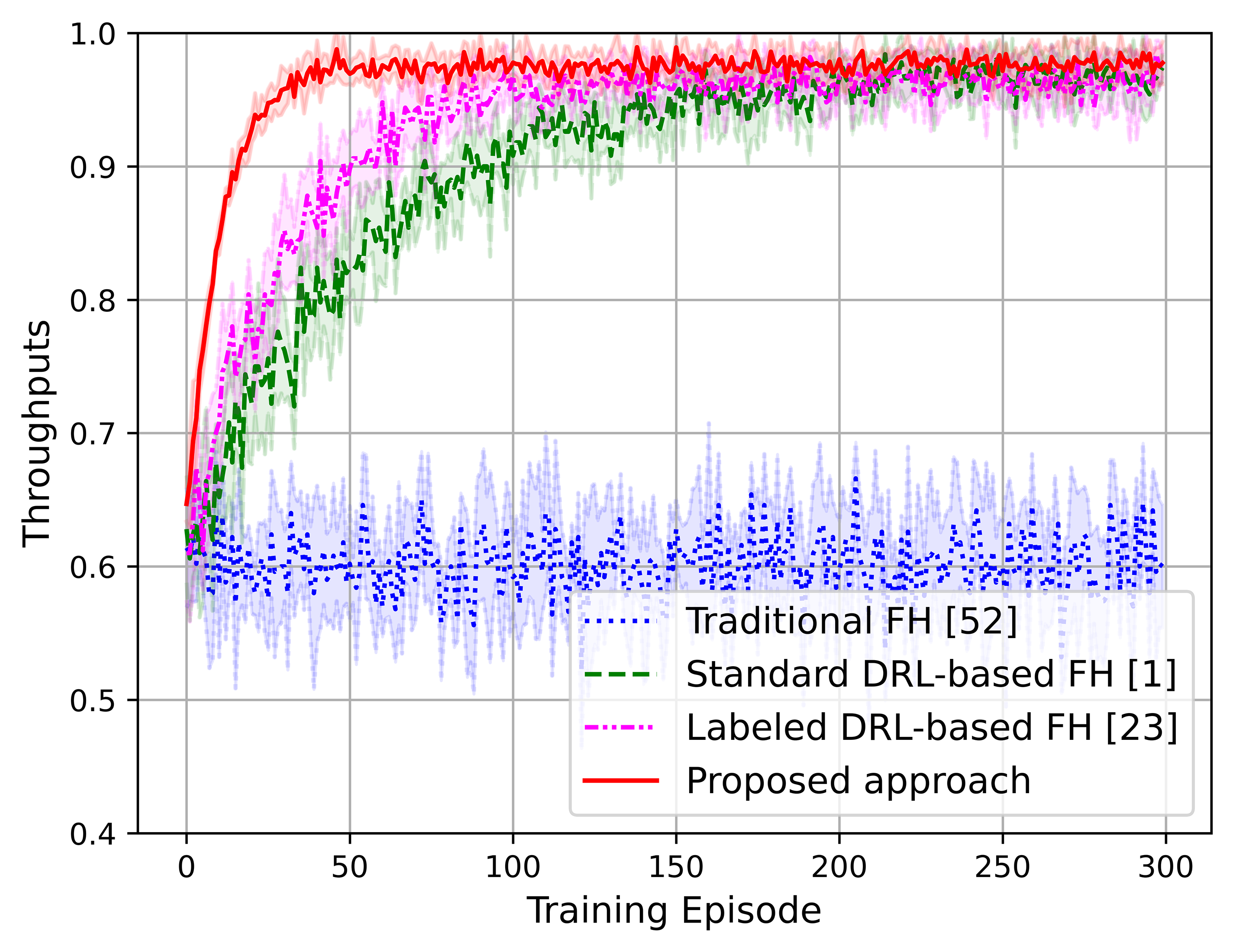}%
\label{8f}}
\caption{Anti-jamming performance of the proposed approach under traditional jamming modes.}
\label{fig_8}
\end{figure*}

We first investigate the performance of the proposed fast adaptive anti-jamming approach under traditional jamming attacks. Specifically, we assume that the intelligent jammer is dormant, while the fixed-mode jammer could launch traditional jamming attacks with a fixed jamming pattern, such as sweeping jamming or comb jamming, etc. In this case, the anti-jamming problem could be simplified to a Markov decision process. The proposed approach is compared with the traditional pseudo-random sequence based FH approach \cite{4215896}, the emerging DRL-based FH approach \cite{8999433}, and the state-of-the-art (SOTA) DRL-based approach (labeled DRL based anti-jamming approach \cite{10227374}), with the results shown in Fig. \ref{fig_8}. In our simulations, we employ a Rayleigh fading channel model to capture the time-varying characteristics of wireless channels in non-line-of-sight (NLOS) scenarios. The Rayleigh distribution could effectively characterize the statistical fluctuations of the received signal envelope in such environments, thereby providing a more accurate representation of realistic fading behavior and introducing appropriate variability in the received signal power over time. As mentioned in \cite{9105045}, the anti-jamming performance of the legitimate user could be evaluated using the normalized throughput, which is defined as the successful transmission rate in every $K$ time slots, i.e.,
\begin{equation}
\mathcal{T} = \frac{1}{K}\sum_{k=1}^{K} \delta( \beta(k, f_k^u) \geq \beta_\text{th} ), \tag{34}\label{34}
\end{equation}
where $\beta(k, f_u^k)$ represents the SINR at the receiver during the $k$-th time slot, $\beta_\text{th}$ denotes the demodulation threshold, and $\delta(\cdot)$ is the indicator function. It is noted that each curve is averaged over 5 independent trials. Across different jamming modes, the throughput of the pseudo-random sequence-based approach remains around 65\%, whereas both the DRL-based and labeled DRL-based approaches converge to approximately 97\%. This occurs because, under fixed-mode jamming attacks, the RL-based anti-jamming approach could effectively learn the jammer's behavior and generate corresponding anti-jamming strategies. Moreover, the proposed approach achieves a throughput close to 100\% and demonstrates a substantially faster rate of convergence during model training compared to the SOTA DRL-based approach, reducing the required training episodes by approximately 80\%. This improvement is attributed to the additional information gain that the coarse-grained spectrum prediction provides to the DRL.

Then, we compare the proposed fast adaptive anti-jamming approach with several existing approaches under the combined influence of a fixed-mode jammer and a DRL-based jammer. The fixed-mode jammer could launch sweeping jamming signals \cite{9733393} with a sweeping speed of 500MHz/s, effectively covering the entire communication band through linear frequency changes. In the experiment, the following five comparison approaches are considered:
\begin{enumerate}[label=\textbullet]
\item{Traditional FH \cite{4215896}: The working frequency of the user is determined by a predefined FH pattern with finite length.}
\item{Standard DRL-based approach \cite{8999433}: The jammer is regarded as a component of the anti-jamming environment, and the DQN based approach could be introduced for combating jamming attacks. Specifically, the network hyperparameters of the DRL-based anti-jamming approach are set according to \cite{8314744}.}
\item{Labeled DRL-based FH \cite{10227374}: A SOTA DRL-based frequency hopping approach that employs soft labels rather than rewards to accelerate the rate of convergence during the training of the anti-jamming model.}
\item{NE-based approach \cite{9292435}: The minimax Q network is used to determine the NE strategies for both the legitimate user and the intelligent jammer.}
\item{Opponent modeling based FH \cite{9939159}: This method adopts the minimax DQN to approximate the user's utility while the imitation learning model is employed to infer the intelligent jammer's strategy. The OMAIJ algorithm could adapt to the dynamic behavior of the intelligent jammer and identify the BR instead of aiming for the NE.}
\end{enumerate}

\begin{figure}[!htb]
\centering
\includegraphics[width=2.5in]{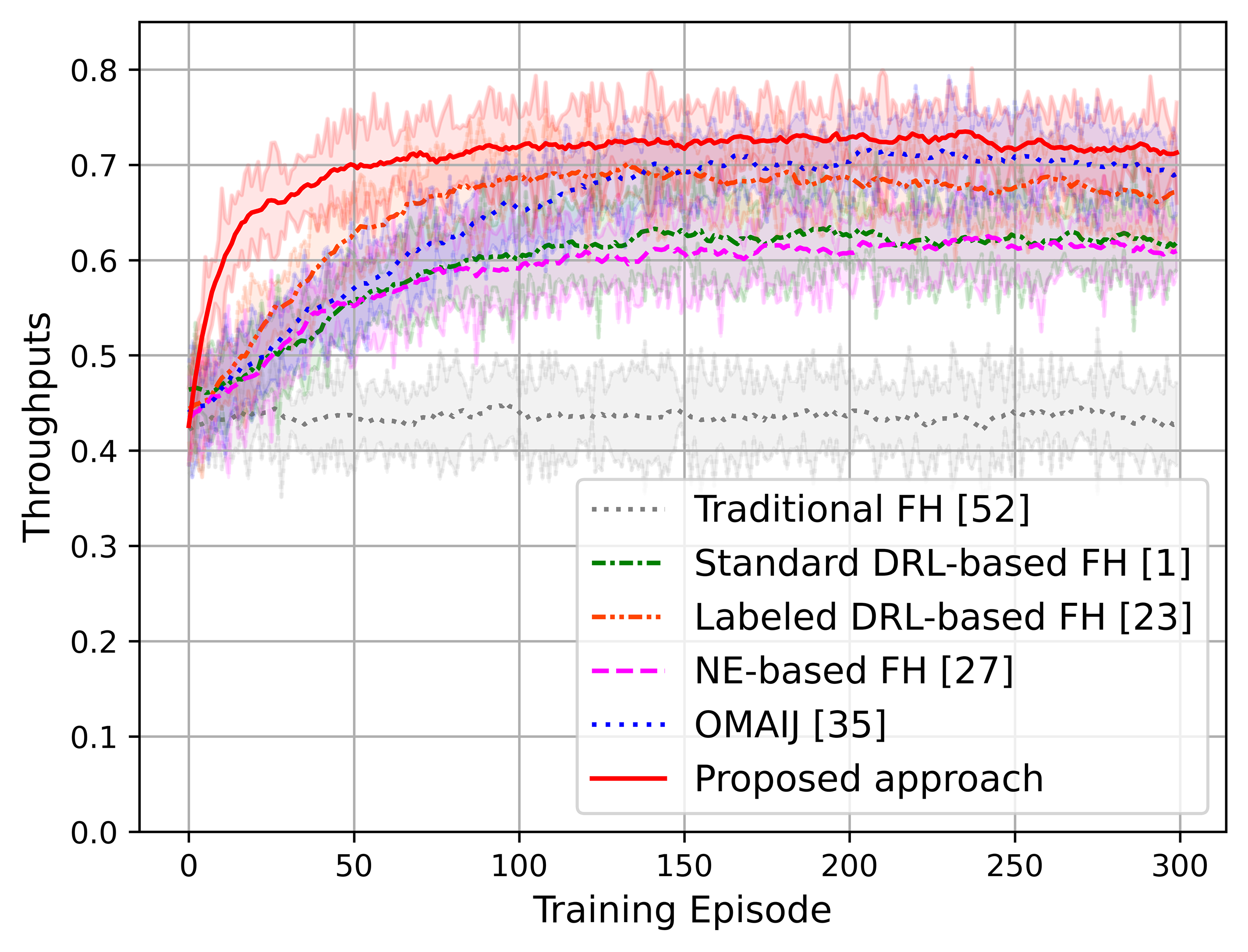}
\caption{Comparison of different anti-jamming approaches under DRL-based intelligent jamming attacks.}
\label{fig_9}
\end{figure}

The normalized throughputs of these anti-jamming approaches over 300 training episodes are shown in Fig. \ref{fig_9}, where each curve is averaged over 5 independent trials. It is evident that our approach outperforms the traditional FH approach, and could achieve about 10\% improvements over the NE-based approach and the standard DRL-based approach. The traditional frequency hopping approach makes it difficult for the RL-based jammer to devise an effective response, but it does not enhance the legitimate user’s anti-jamming performance. What's more, the learning based anti-jamming algorithms could increase the user's normalized throughput by learning the jammers' dynamics, and we find that the anti-jamming performance of the NE-based and standard DRL-based approaches is comparable. Compared to SOTA DRL-based frequency hopping approach \cite{10227374}, our approach demonstrates a slight improvement in anti-jamming performance and reduces the number of training episodes by approximately 50\%. The proposed approach could achieve comparable anti-jamming performance to the opponent modeling based approach. Meanwhile, the proposed approach also demonstrates a significant acceleration in the rate of convergence and could reduce the number of training episodes by up to 70\%. Moreover, our approach adopts the coarse-grained spectrum prediction as the auxiliary task and does not need any knowledge of the jamming action or the jammer's action space, which might not be accessible in practical anti-jamming scenarios.

\begin{figure}[!htb]
\centering
\includegraphics[width=2.7in]{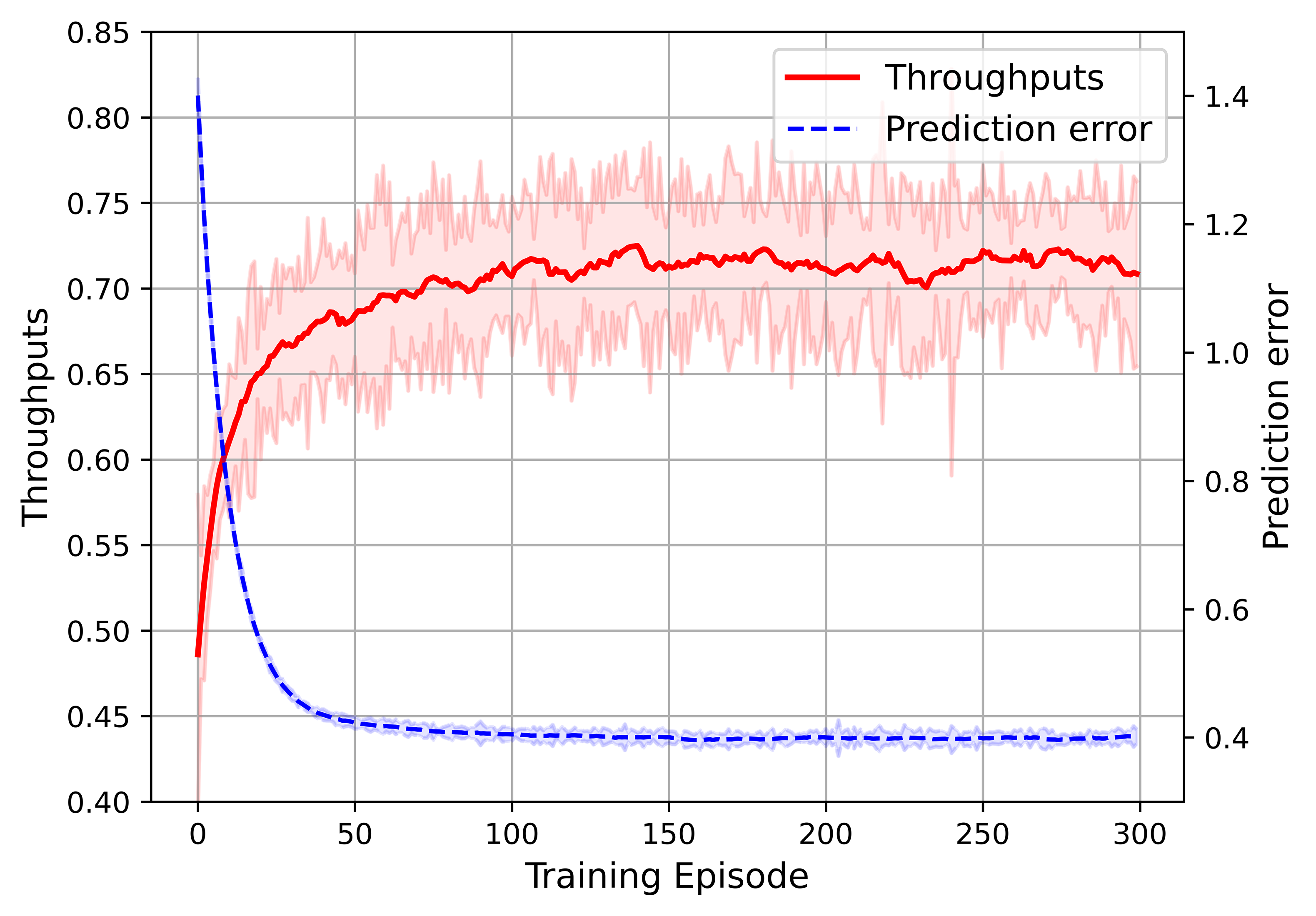}
\caption{The relationship between coarse-grained spectrum prediction error and anti-jamming performance during the online training process.}
\label{fig_10}
\end{figure}

Additionally, we use prediction error to quantify the accuracy of the coarse-grained spectrum prediction and further investigate the accuracy of this CNN-based prediction during the online training process. Specifically, the coarse-grained spectrum prediction error is defined as
\begin{equation}
D(\hat{\bf{c}}_n, {\bf{c}}_n) = \frac{\sqrt{\frac{1}{M} \sum_{m=1}^M \left ( \hat{c}_n^m - c_n^m \right )^2}}{\max \left \{{\bf{c}}_n \right \} - \min \left \{ {\bf{c}}_n \right \}}, \tag{35}\label{35}
\end{equation}
where $\hat{c}_n^m$ denotes the $m$-th element of the predicted coarse-grained spectrum $\hat{\bf{c}}_n$, and $c_n^m$ denotes the $m$-th element of the ground-truth label ${\bf{c}}_n$. Fig. \ref{fig_10} shows the relationship between coarse-grained spectrum prediction error and anti-jamming performance during the online training process. It can be observed that the coarse-grained spectrum prediction error gradually decreases with the increasing number of training episodes, while the throughput of the legitimate user correspondingly improves. This is because the accurately-estimated coarse-grained spectra assist the joint decision function in identifying better anti-jamming strategies, thereby significantly improving the performance of the anti-jamming agent.

\begin{figure}[!htb]
\centering
\includegraphics[width=2.5in]{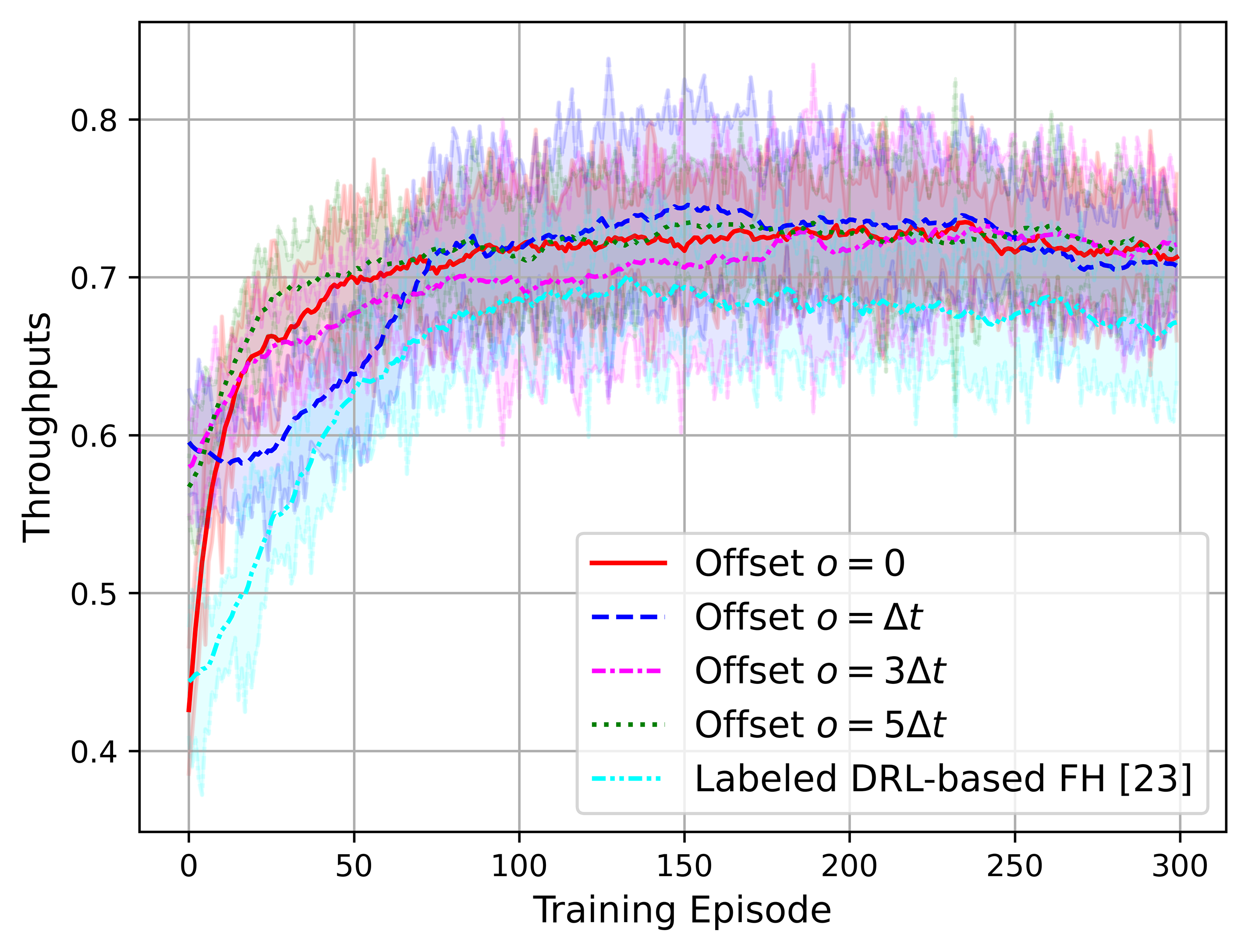}
\caption{The performance of the proposed fast-adaptive anti-jamming method with non-synchronous time-slotted framing structure.}
\label{fig_11}
\end{figure}

\begin{figure}[!htb]
\centering
\includegraphics[width=2.5in]{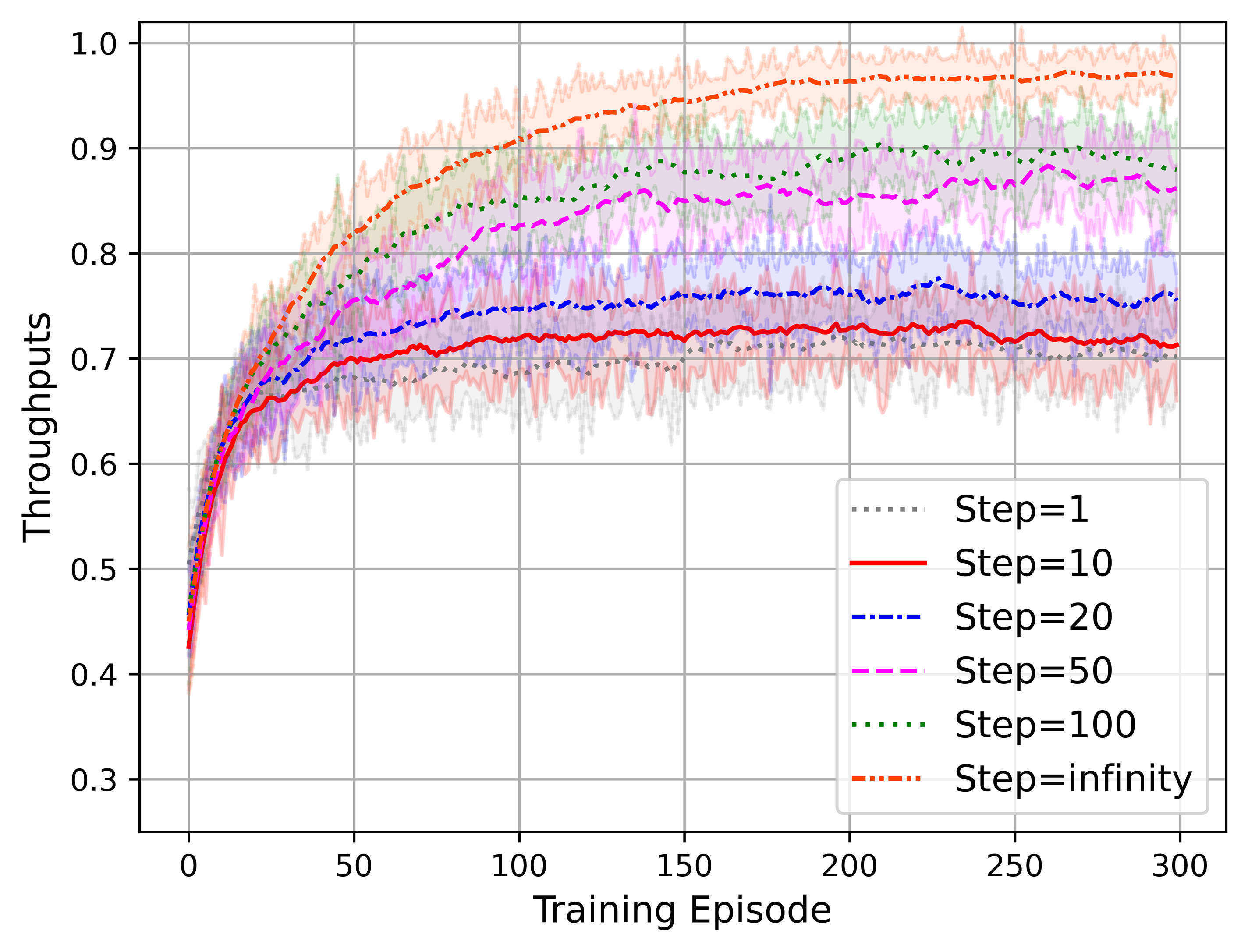}
\caption{Comparison of the proposed approach under different jamming update steps.}
\label{fig_12}
\end{figure}

Furthermore, we investigate the anti-jamming performance of the proposed fast adaptive anti-jamming approach under a more realistic non-synchronous time-slotted framing structure through simulations, where an offset $o \in \left ( 0, T_h/2 \right )$ is assumed due to the switching of anti-jamming actions and jamming actions. The results are shown in Fig. \ref{fig_11}, where the offset is set to $o = \Delta t, 3\Delta t, 5\Delta t$, respectively. It can be seen that the misalignment of time slots has no significant impact on the performance of the proposed anti-jamming approach. Even with an offset of $o =5\Delta t$, the anti-jamming performance of the proposed approach still significantly outperforms the SOTA DRL-based anti-jamming approach \cite{10227374}, which uses a synchronous time-slotted framing structure, thereby demonstrating the robustness of the proposed fast adaptive anti-jamming approach. Additionally, we also investigate the anti-jamming performance of the proposed approach when facing the intelligent jammer with different update steps, and the results are shown in Fig. \ref{fig_12}. The smaller update step implies that the jammer's policy could be updated more frequently. As the jammer accelerates its updating process, the anti-jamming performance deteriorates. This is because both the coarse-grained spectrum prediction network and the Q-function estimation network find it more challenging to learn the rapidly changing jamming policy. But even if the intelligent jammer takes the same update frequency as the legitimate user (i.e., Step=1), the normalized throughput could still achieve 70\%, which is also much higher than the NE-based approach. If the intelligent jammer adopts a fixed jamming policy (i.e., Step = infinity), the normalized throughput could converge to about 95\%. Therefore, if the user could update its policy more rapidly, the proposed approach could perform better when facing learning-based jammers.

\begin{figure}[!htb]
\centering
\includegraphics[width=2.5in]{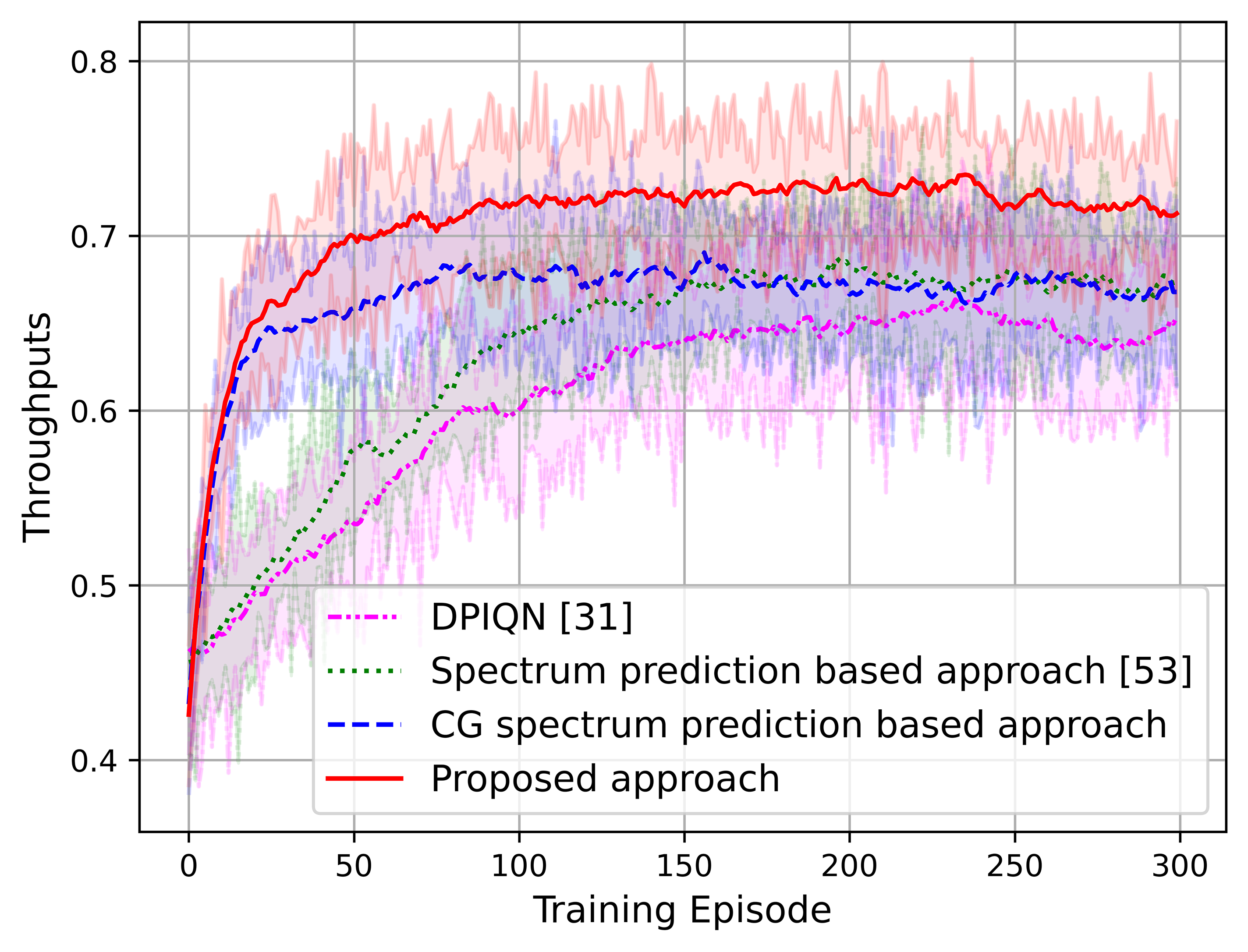}
\caption{Ablation experiment.}
\label{fig_13}
\end{figure}

Finally, we conducted an ablation experiment to validate the effectiveness of the proposed approach. As shown in Fig. \ref{fig_13}, the proposed approach is compared with the opponent modeling based approach (DPIQN) \cite{20184205945705}, the coarse-grained spectrum prediction based approach, and the general spectrum prediction based approach \cite{10039050}. The DPIQN algorithm also utilizes an auxiliary task for the opponent's policy feature extraction and selects the action with the maximum Q-value in each hop, while the spectrum prediction based approaches select the channel with the minimum predicted discrete spectrum sample value for data transmission at the beginning of each hop. The results in Fig. \ref{fig_13} demonstrate that the proposed approach exhibits the strongest anti-jamming performance and the fastest rate of convergence during model training, followed by the supervised learning based coarse-grained spectrum prediction approach and the general spectrum prediction based approach, with the DPIQN based Q-function estimation approach being the least effective. This indicates that the supervised learning based approach converges faster than the DPIQN algorithm, and the proposed approach could effectively leverage the features extracted from DRL, as well as coarse-grained spectrum prediction, for joint decision-making. It is noted that the proposed approach and the coarse-grained spectrum prediction based approach demonstrate a reduction of up to 70\% in training episodes compared to the general spectrum prediction based approach. This is because, compared to the general spectrum prediction, the proposed coarse-grained spectrum prediction task is much simpler. This improvement stems from the simplicity of the proposed coarse-grained spectrum prediction task compared to general spectrum prediction.

\subsection{Computational Complexity Analysis}
The computational complexity for the proposed fast adaptive anti-jamming approach is measured in floating-point operations (FLOPs). In the case of a CNN, the computational complexity of a Conv layer is defined as
\begin{equation}
\text{FLOPs}_\text{Conv} = 2 C_{\text{in}}K^2 \times (H \times W \times C_{\text{out}}), \tag{36}\label{36}
\end{equation}
where $K$ is the kernel size, $H \times W$ is the size of the output feature map, $C_{\text{in}}$ and $C_{\text{out}}$ are the number of input channels and output channels, respectively. 

The computational complexity of a FC layer is defined as
\begin{equation}
\text{FLOPs}_{\text{FC}} = 2I \times O, \tag{37}\label{37}
\end{equation}
where $I$ and $O$ are the number of input and output neurons.

The required FLOPs for both the Q-function estimation network and the coarse-grained spectrum prediction network are $143.9 \times 10^6$, and the complexity of the proposed Algorithm 1 is $287.8 \times 10^6$. Graphics processing unit (GPU) could satisfy the computational demands of the proposed anti-jamming algorithm with joint DQN and coarse-grained spectrum prediction. In simulations, each training episode—including all channel sensing, access, and jamming operations—takes approximately 5 seconds when executed on an NVIDIA RTX 4090 GPU. The actual duration may vary depending on the specific hardware platform and the simulation configuration. Additionally, a comparison between the proposed fast adaptive anti-jamming channel access approach and several benchmarks in terms of computational complexity, rate of convergence, and anti-jamming performance is shown in Table \ref{tab2}. It can be observed that although the proposed approach incurs higher computational complexity than the standard DRL-based method \cite{8999433}, labeled DRL-based method \cite{10227374}, and NE-based method \cite{9292435}, due to the use of dual-branch CNNs and coarse-grained spectrum prediction, it achieves significantly higher throughput and a faster rate of convergence during training. Furthermore, compared to the opponent modeling-based method (i.e., OMAIJ \cite{9939159}) with similar computational complexity, our approach attains slightly better anti-jamming performance and converges more rapidly. Therefore, the tradeoff between complexity and performance is justified by the improved anti-jamming effectiveness and training efficiency.

\begin{table*}[!htb]
\caption{Comparison of computational complexity, rate of convergence, and anti-jamming performance}
\label{tab2}
\centering
\begin{tabular}{|c|c|c|c|}
\hline
Method & \makecell{Computational Complexity \\ (FLOPs)} & \makecell{Rate of convergence \\ (episodes)} & Throughput \\ \hline
Proposed approach & 287.8$\times10^6$ & 50 & 0.73 \\ \hline
OMAIJ \cite{9939159}           & 287.3$\times10^6$ & 150 & 0.70 \\ \hline
Standard DRL-based FH \cite{8999433}     & 143.6$\times10^6$ & 150 & 0.62 \\ \hline
NE-based FH \cite{9292435}  & 143.6$\times10^6$ & 120 & 0.61 \\ \hline
Labeled DRL-based FH \cite{10227374}     & 48.9$\times10^6$ & 100 & 0.66 \\ \hline
\end{tabular}
\end{table*}

\section{Conclusion}
This paper proposes a novel fast adaptive anti-jamming channel access approach that leverages deep Q learning and CG spectrum prediction to enhance anti-jamming performance while reducing training episodes in wireless confrontations. We formulate the adversarial interactions as an MG and address the limitations of existing opponent modeling techniques that rely on the observation of opponent actions by employing a supervised learning based CG spectrum prediction. This prediction serves as an auxiliary task for learning the intelligent jammer's strategy, and an updated training dataset employing the FIFO principle is adopted to ensure its responsiveness to dynamic environments. Numerical simulations indicate that the proposed approach demonstrates a reduction of up to 70\% in training episodes compared to DRL-based approaches, along with a 10\% improvement in throughput over NE strategies. This work contributes a robust, efficient solution for adaptive anti-jamming in complex, real-world scenarios. In the future, it would be valuable to explore practical implementations of the proposed anti-jamming approach, which could perform joint DQN and CG spectrum prediction, using software-defined radio devices.

\bibliographystyle{unsrtnat}
\bibliography{ref.bib}  






\end{document}